\documentclass[pre,%
 reprint,
 amsmath,amssymb,
 aps,
floatfix,
]{revtex4-2}
\usepackage{placeins}
\usepackage{makecell}
\usepackage{graphicx}
\usepackage{dcolumn}
\usepackage{bm}
\usepackage{subfigure}
\usepackage{amsmath}
\usepackage{algpseudocode}
\usepackage{xcolor}
\usepackage{bm}
\usepackage{booktabs}  

\newcommand{\trainingTimesX}{4}

\newcommand{\percentageOne}{96.6}
\newcommand{\percentageTwo}{96.4}

\newcommand{\percentageThree}{90.68}
\newcommand{\percentageFour}{71.62} 
\newcommand{\lambdaEigenOptimal}{100}
\newcommand{\lambdaSylvesterOptimal}{100}

\newcommand{\percentageFive}{40.87}
\newcommand{\percentageSix}{97.55}

\newcommand{\avgLossGeLU}{0.007}
\newcommand{\avgLossXTanhKX}{0.011}
\newcommand{\avgLossSoftplus}{0.144}
\newcommand{\avgLossHybrid}{0.168}

\newcommand{\lOneLossOurs}{0.01523}
\newcommand{\lOneLossOriginal}{0.073}
\newcommand{\percentageSeven}{79.1}

\newcommand{\energyErrorPercentage}{3.6}

\newcommand{\validationLossTripleStart}{1.38029}
\newcommand{\validationLossTriple}{0.019724}
\newcommand{\energyErrorNonChaotic}{0.58}

\newcommand{\percentageAbstractOne}{96.6}

\newcommand{\percentageAbstractTwo}{90.68}

\newcommand{\figRegularizationExistence}{figures/regularization_existence.pdf}
\newcommand{\capRegularizationExistence}{Comparison of training loss with and without regularization. Both eigenvalue and Sylvester's condition regularization methods enable successful training where unregularized LNN are unstable and fail more often.}

\newcommand{\figRegularizationComparison}{figures/regularization_comparison.pdf}
\newcommand{\capRegularizationComparison}{Training stability comparison between baseline models, eigenvalue regularization and Sylvester's condition regularization.}

\newcommand{\figLambdaValues}{figures/lambda_values.pdf}
\newcommand{\capLambdaValues}{Minimum loss comparison between eigenvalue and Sylvester regularization methods. Both methods achieve similar final performance. Optimal values are $\lambda = \lambdaEigenOptimal$ for eigenvalue and $\lambda = \lambdaSylvesterOptimal$ for Sylvester regularization.}

\newcommand{\figScalingPerformance}{figures/scaling_performance.pdf}
\newcommand{\capScalingPerformance}{Performance comparison of physical scaling vs. no scaling vs. DyT normalization.}

\newcommand{\figActivationComparison}{figures/activation_comparison.pdf}
\newcommand{\capActivationComparison}{Comparison of different activation functions. GeLU achieves best performance, followed by $x\tanh(kx)$.}

\newcommand{\figTrainingRange}{figures/training_range.pdf}
\newcommand{\capTrainingRange}{Comparison of learned mass matrix behavior of a double pendulum across different training ranges. The $[0, 2\pi]$ range exhibits problematic determinant signs near boundaries (left), while the extended $[-\pi/2, 2\pi + \pi/2]$ range exhibits similar shape to the analytical Hessian and produces correct behavior throughout the domain (right). An analytical determinant of the Hessian matrix is provided at the center.}

\newcommand{\figExtraTerm}{figures/extraterm_performance.pdf}
\newcommand{\capExtraTerm}{Trade-off between validation loss and training instability, with and without an initial velocity term.}

\newcommand{\figOldModels}{figures/old_model_comparison.pdf}
\newcommand{\capOldModels}{Performance vs Training Instability Trade-off between two different proposed models in the literature.}

\newcommand{\figDoublePendulumCoordError}{figures/double_pendulum_coord_error.png}
\newcommand{\capDoublePendulumCoordError}{Coordinate prediction error over time for a randomly initialized double pendulum system.}

\newcommand{\figDoublePendulumEnergy}{figures/double_pendulum_energy.png}
\newcommand{\capDoublePendulumEnergy}{Normalized energy error over time for the tested double pendulum system.}

\newcommand{\figDoublePendulumLagrangian}{figures/double_pendulum_lagrangian.png}
\newcommand{\capDoublePendulumLagrangian}{Comparison of network-learned Lagrangian with the analytical Lagrangian of the double pendulum system.}

\newcommand{\figDoublePendulumHessian}{figures/double_pendulum_hessian.png}
\newcommand{\capDoublePendulumHessian}{Comparison of the determinant of network-learned Hessian of the Lagrangian with the determinant of analytical Hessian of the Lagrangian, for the double pendulum system.}

\newcommand{\figDoublePendulumTrace}{figures/double_pendulum_trace.png}
\newcommand{\capDoublePendulumTrace}{Analytical and network-learned trace of the tested double pendulum system over time with density gradient.}

\newcommand{\figSpringPendulumEnergyError}{figures/spring_pendulum_energy_error.png}
\newcommand{\capSpringPendulumEnergyError}{Energy conversation error for 4 randomly initialized spring pendulum systems.}

\newcommand{\figSpringPendulumTrace}{figures/spring_pendulum_trace.png}
\newcommand{\capSpringPendulumTrace}{Comparison of analytical and predicted trajectory traces of the spring pendulum system.}

\newcommand{\figTriplePendulumCoordErrorNonChaotic}{figures/triple_pendulum_coord_error_nonchaotic.png}
\newcommand{\capTriplePendulumCoordErrorNonChaotic}{Position prediction error for a non-chaotic triple pendulum trajectory.}

\newcommand{\figTriplePendulumEnergyNonChaotic}{figures/triple_pendulum_energy_nonchaotic.png} 
\newcommand{\capTriplePendulumEnergyNonChaotic}{Energy conservation over time for non-chaotic triple pendulum trajectory. (Average error: \energyErrorNonChaotic\%)}

\newcommand{\figTriplePendulumCoordErrorChaotic}{figures/triple_pendulum_coord_error_chaotic.png}
\newcommand{\capTriplePendulumCoordErrorChaotic}{Coordinate predictions vs. ground truth for a chaotic triple pendulum trajectory.}

\newcommand{\figTriplePendulumEnergyChaotic}{figures/triple_pendulum_energy_chaotic.png}
\newcommand{\capTriplePendulumEnergyChaotic}{Energy conservation over time for a chaotic triple pendulum trajectory (error: 19.08\%).}

\newcommand{\figTriplePendulumTraining}{figures/triple_pendulum_training.pdf}
\newcommand{\capTriplePendulumTraining}{Training loss curve for the triple pendulum system.}

\newcommand{\figSphereCoordAnalytical}{figures/sphere_geodesic_analytical_coords.png}
\newcommand{\capSphereCoordAnalytical}{The traced path of tested analytically solved geodesic on a sphere.}

\newcommand{\figSphereCoordLearned}{figures/sphere_geodesic_lnn_coords.png}
\newcommand{\capSphereCoordLearned}{The traced path of tested network-learned geodesic on a sphere.}

\newcommand{\figSphereEnergy}{figures/sphere_geodesic_energy_error.png} 
\newcommand{\capSphereEnergy}{Comparison of energy conservation per time for an analytical and a network-learned geodesic on a sphere.}

\newcommand{\figAdSCoordError}{figures/ads4_coord_error.pdf}
\newcommand{\capAdSCoordError}{Traced learned and analytically solved path for geodesics in AdS4 spacetime.}

\newcommand{\figAdSTDot}{figures/ads4_tdot.pdf}
\newcommand{\capAdSTDot}{$\dot{t}$ (time dilation) comparison between an analytical and a network-learned geodesic in AdS4 spacetime.}

\newcommand{\figAdSZDot}{figures/ads4_zdot.pdf}
\newcommand{\capAdSZDot}{$\dot{z}$ comparison between an analytical and a network-learned geodesic in AdS4 spacetime.}

\newcommand{\capActivationTable}{Comparison of validation loss and training instability across different activation functions. GeLU achieved the best overall performance with lowest validation loss and instability, while $xtanh(kx)$ followed closely. The Softplus and hybrid approaches exhibited substantially higher validation loss, indicating poorer generalization.}

\newcommand{\capInitializationTable}{Performance comparison of custom vs. default weight initialization method across different activation functions. Custom initialization improves Softplus and $x\tanh(kx)$ but slightly degrades GeLU performance.}

\begin{document}

\preprint{APS/123-QED}

\title{Learning Relativistic Geodesics and Chaotic Dynamics via Stabilized Lagrangian Neural Networks}

\author{Abdullah Umut Hamzaogullari$^{1,2,*}$}
\author{Arkadas Ozakin$^{1}$}

\affiliation{
$^{1}$Department of Physics, Bogazici University, Istanbul, Turkey\\
$^{2}$Department of Computer Engineering, Bogazici University, Istanbul, Turkey\\
$^{*}$abdullah.hamzaogullari@std.bogazici.edu.tr
}

\newcommand{\zm}[1]{{\color{black!0!blue} #1}}

\date{\today}

\begin{abstract}
Lagrangian Neural Networks (LNNs) can learn arbitrary Lagrangians from trajectory data, but their unusual optimization objective leads to significant training instabilities that limit their application to complex systems. We propose several improvements that address these fundamental challenges, namely, 
a Hessian regularization scheme that penalizes unphysical signatures in the Lagrangian's second derivatives with respect to velocities, preventing the network from learning unstable dynamics,
activation functions that are better suited to the problem of 
learning Lagrangians, and a physics-aware coordinate scaling
that improves stability. We systematically evaluate these techniques alongside previously proposed methods for improving stability. Our improved architecture successfully trains on systems of unprecedented complexity, including triple pendulums, and achieved \percentageAbstractOne{}\% lower validation loss value and \percentageAbstractTwo{}\% better stability than baseline LNNs in double pendulum systems. With the improved framework, we show that our LNNs can learn
Lagrangians representing geodesic motion in both
non-relativistic and general relativistic settings.
To deal with the relativistic setting, we extended our regularization to penalize violations of Lorentzian signatures, which allowed us to predict a geodesic Lagrangian under AdS\textsubscript{4} spacetime metric directly from trajectory data, 
which to our knowledge has not been done in the literature before. This opens new possibilities for automated discovery of geometric structures in physics, including extraction of spacetime metric tensor components from geodesic trajectories. While our approach inherits some limitations of the original LNN framework, particularly the requirement for invertible Hessians, it significantly expands the practical applicability of LNNs for scientific discovery tasks. The code is available at https://github.com/abdullahumuth/hessian\_regularized\_lnn.
\end{abstract}

\maketitle


\section{Introduction} \label{sec:intro}
Neural network-based models of physical systems learn to predict
the behavior of a system directly from data. In the most straightforward approach,
the network has no built-in structure that enforces physicality constraints, 
and as a result, even when the model does a reasonable job in predicting trajectories,
it may fail to preserve physical quantities such as energy over long periods of 
time. Lagrangian Neural Networks (LNNs) \citep{cranmerLagrangianNeuralNetworks2020} 
overcome this problem through a physics-based architecture which works by
learning to predict the Lagrangian function of the system, and doing trajectory 
computation via this learned Lagrangian. Combining the 
capability of LNNs to learn Lagrangians from trajectory data with
a symbolic learning approach makes it possible to pursue
autonomous scientific discovery \citep{NEURIPS2020_c9f2f917}.

However, due to their unusual learning objective \citep{cranmerLagrangianNeuralNetworks2020}, the 
training of LNNs can be unstable \citep{Liu:2021xwt}. Unlike standard neural networks, LNNs must compute second derivatives and invert the resulting Hessian matrix during training, which makes optimization particularly challenging. The training of LNNs for complex systems
is particularly susceptible to various problems,
and custom special initializations \citep{cranmerLagrangianNeuralNetworks2020} or other training tricks \citep{finziSimplifyingHamiltonianLagrangian2020,Liu:2021xwt}
have been developed to help improve the methodology. Despite
these efforts, it has been challenging to get good results 
for complicated systems.

We present a modified LNN architecture with a Hessian regularization term 
added to the loss function, as well as other critical changes including 
improved activation functions and physics-aware scaling, to make the 
system more stable and more powerful. We then use the proposed 
architecture to train complicated mechanical systems, including triple 
pendulums. As a more novel demonstration of the capabilities 
our approach, we use the proposed model to learn Lagrangians representing 
classical geodesic motion on curved surfaces and
geodesic motion in general relativistic spacetimes, with 
spacetime coordinates being used as inputs in the latter case.

The paper is organized as follows: Section~\ref{sec:relatedwork} reviews related work, Section~\ref{sec:methods} presents our methods, Section~\ref{sec:networkexp} shows the improvements the proposed changes to the network have on the results, Section~\ref{sec:physicalexp} shows physical system experimental results, and Section~\ref{sec:conclusion} concludes with discussion and future directions.

\section{Related Work} \label{sec:relatedwork}
Tips and tricks to boost training of LNNs start to appear with the original LNN paper \cite{cranmerLagrangianNeuralNetworks2020}. The authors argue that LNNs have an unusual optimization objective and while Kaiming \citep{He:2015dtg} and Xavier \citep{glorotUnderstandingDifficultyTraining2010} initializations are optimized for regular neural networks, they are not sufficient to train an LNN successfully. 
They propose a custom initialization scheme for LNNs
by empirically optimizing the KL-divergence of the gradient of each parameter with respect to a univariate Gaussian. 
They then perform a symbolic regression to predict the optimal initialization variance for varying values of neural network depth and width, but for specific input sizes. 
This approach requires re-optimizing the initialization for each input size, which may limit its practicality, and is not demonstrated to suffice for higher dimensional systems.

\citet{Liu:2021xwt} proposed two simpler tricks to boost training than custom initializations.
They claim that Softplus, the activation function that is used in \citep{cranmerLagrangianNeuralNetworks2020}, is general but inefficient 
for representing quadratic functions easily. Since
quadratic functions are commonly encountered in physics,
the authors propose to divide the network into two pieces,
where one piece uses Softplus as the activation function and 
the other piece uses the quadratic function. 
The two networks output scalars 
($\mathcal{L}_1$ and $\mathcal{L}_2$), which are
added to form the Lagrangian. The other trick they propose is to add an initial $\dot{\mathbf{q}}^T\dot{\mathbf{q}}$ term to the Lagrangian, so that the total Lagrangian becomes
$$\mathcal{L} = \mathcal{L}_1 + \mathcal{L}_2 + \alpha \dot{\mathbf{q}}^T\dot{\mathbf{q}}$$
with $\alpha$ chosen to be 1.
The initial $\dot{\mathbf{q}}^T\dot{\mathbf{q}}$ term will ensure that the Hessian, or the mass matrix, will start off as a positive definite matrix, which will always be invertible.

\citet{finziSimplifyingHamiltonianLagrangian2020} propose Constrained Hamiltonian Neural Networks (CHNNs) and Constrained Lagrangian Neural Networks (CLNNs) which modify the HNN \citep{greydanusHamiltonianNeuralNetworks2019} and LNN architectures by baking explicit constraints into the model and always training them in cartesian coordinates. 
This approach greatly reduces the data needed to train, the training scenarios converge faster, and the results are more accurate than HNNs, NeuralODEs \citep{chenNeuralOrdinaryDifferential2018} and DeLaNs \citep{Lutter:2019ovq}, although there is no direct comparison with an LNN. 
A key stability advantage of CLNNs is that they naturally avoid coordinate singularities in 3D systems (gimbal locks) which happen if the generalized coordinates are given by Euler angles \citep{finziSimplifyingHamiltonianLagrangian2020}.
One of the limitations of this approach is that together with trajectory data, it requires knowing the system's explicit constraints beforehand, which makes the architecture preferable to authors looking to solve practical problems in robotics and control-related tasks, like \citet{whiteStabilizedNeuralDifferential2023}, instead of theoretical problems such as finding new physics \citep{Liu:2021xwt} or scientific discovery \citep{NEURIPS2020_c9f2f917}.

\citet{fuTwoAIScientists2025} propose MASS, which is inspired by HNNs and LNNs and has some of the same training challenges, such as calculating Hessian inverses and matrix inverses that consists of other second derivatives. 
The authors claim that under Kaiming \citep{He:2015dtg} and Xavier \citep{glorotUnderstandingDifficultyTraining2010} initializations, the second derivatives are very small, thus resulting in exploding inverses. 
Because of this, they augment their inputs from $(\mathbf{x},\mathbf{y})$ to $(\mathbf{x},\mathbf{y},\dot{\mathbf{x}},\dot{\mathbf{y}},\mathbf{x}\mathbf{y})$, which make the training of MASS networks stable with learning rates up to $10^{-2}$.

\section{Methods} \label{sec:methods}

While we tackle the training challenges of LNNs, the goal is to maintain their core advantages. Our approach follows several key principles: (i) We preserve the original LNN framework's ability to learn arbitrary physically valid Lagrangians without restricting functional forms. (ii) Like the original LNN, we require only coordinate trajectory data, with no additional information about forces or system parameters. (iii) The method remains applicable to black-box systems where we assume no prior knowledge except for the angular periodicity of certain coordinates. (iv) Our improvements generalize to more complex systems without requiring specialized tuning for each case.

These principles ensure that the improved architecture remains suitable for scientific discovery tasks while solving the most common stability problems.
\subsection{Hessian Regularization} \label{subsec:hessian}

The primary source of training instability in LNNs stems from the computation of the inverse of the Lagrangian's Hessian with respect to generalized velocities, $\frac{\partial^2 \mathcal{L}}{\partial \dot{\mathbf{q}}^2}$, commonly referred to as the mass matrix of the system.
This inverse is required to compute accelerations through the Euler-Lagrange equations:
\begin{equation}
\ddot{\mathbf{q}} = (\nabla_{\dot{\mathbf{q}}}\nabla_{\dot{\mathbf{q}}}^{\top}\mathcal{L})^{-1}[\nabla_{\mathbf{q}} \mathcal{L} - (\nabla_{\mathbf{q}}\nabla_{\dot{\mathbf{q}}}^{\top}\mathcal{L})\dot{\mathbf{q}}]
\label{eq:acceleration}
\end{equation}
The usual LNN loss function is:
\begin{equation}
\mathcal{L}_{\text{usual}} = \text{MAE}(\ddot{\mathbf{q}}_{\text{real}}, \ddot{\mathbf{q}}_{\text{predicted}})
\label{eq:usual_loss}
\end{equation}

During training, if the determinant of the mass matrix gets close to or equals zero, the inverse terms become arbitrarily large, resulting in destabilizing acceleration values and loss gradients, since the 
gradient of the loss is also dependent on the Hessian inverse. (For a full derivation of the gradient of the loss, we refer to Appendix \ref{Appendix:A}.)

Note that throughout this subsection, we refer to and give the equations for single particle Lagrangians, although the logic is fully extendable to multi-particle systems.

\textbf{Physical Constraints.} In classical (nonrelativistic)
mechanics, the mass matrix 
of a system can't have negative eigenvalues---systems violating this 
constraint would have kinetic energy terms unbounded below in certain 
directions, producing classical runaway solutions. Such systems are 
fundamentally unstable and at best metastable \citep{Gross:2020tph}. 
For the general
relativistic case of a particle moving along the geodesics of 
a given spacetime, timelike directions \footnote{We adopt the
mostly minus metric signature, i.e.,
$(-1, +, +, +$ in the 4-dimensional case} 
do allow negative eigenvalues
of the (generalized) mass matrix---in this case, one has 
a constraint on the signs of the eigenvalues of the matrix
that allows only one eigenvalue to be negative.

Below, we formulate a regularization approach to enforce these
two types of constraints. However, since the
mass matrix may legitimately become non-invertible in specific physical 
scenarios including gauge freedoms and coordinate 
singularities (e.g., gimbal lock in Euler angle representations), 
our regularization only prevents incorrect signs while allowing zero 
eigenvalues thus allowing future developments that can deal with 
such settings.

\textbf{Proposed Regularization.} We introduce regularization terms that penalize violations of fundamental signature constraints on the mass matrix $M$. We present two approaches: direct eigenvalue penalty for classical systems, and a scale-invariant method based on Sylvester's criterion, configured for both classical systems and relativistic systems with a time-like coordinate.

\textbf{Option 1: Eigenvalue Regularization}

For classical systems where generalized coordinates $\mathbf{q}$ are purely spatial, we directly penalize negative eigenvalues of the mass matrix:
\begin{equation}
\mathcal{L}_{\text{eigen}} = \text{MAE}(\ddot{\mathbf{q}}_{\text{real}}, \ddot{\mathbf{q}}_{\text{predicted}}) + \lambda \sum_{\lambda_i < 0} |\lambda_i|
\label{eq:eigen_loss}
\end{equation}
where $\lambda_i$ are the mass matrix eigenvalues and $\lambda$ is the regularization coefficient. This penalty guides the network away from unphysical Lagrangians. We allow zero eigenvalues to accommodate future extensions that might handle singular mass matrices arising from valid physical constraints.

Regarding computational cost, eigenvalue decomposition requires $O(n^3)$ operations. However, the standard LNN framework already computes the Hessian inverse, which is also an $O(n^3)$ operation. Thus, eigenvalue regularization does not increase the asymptotic complexity of training. Moreover, since coordinate dimensions $n$ in typical physics problems are small (e.g., 2 or 3), the practical overhead remains minimal.

\textbf{Option 2: Scale-Invariant Sylvester's Criterion}

We propose an alternative using Sylvester's criterion, which is more easily generalized to the relativistic case.

We define characteristic scales $c_i$ for each coordinate $i$ as the median of $|M_{ii}|$ across training batches to better work with coordinates that may have values that are orders of magnitude different from one another.
\subsubsection{Classical Systems}
For classical systems, the mass matrix must be positive semi-definite. Sylvester's criterion requires all principal minors to have non-negative determinants. However, checking all principal minors is computationally prohibitive. We must therefore select a subset to evaluate. We choose the leading principal minors, which is sufficient to verify positive definiteness, as they provide a natural choice for our regularization. Our scale-invariant loss becomes:
\begin{equation}
\mathcal{L}_{\text{classical}} = \text{MAE}(\ddot{\mathbf{q}}_{\text{real}}, \ddot{\mathbf{q}}_{\text{predicted}}) + \lambda \sum_{k=1}^{n} \frac{\text{ReLU}(-\det(M_k))}{s_k + \epsilon}
\label{eq:sylvester_classical_loss}
\end{equation}
where $s_k = \prod_{i=1}^{k} c_i$ accounts for the determinant's scaling, and $\epsilon$ ensures numerical stability.

\subsubsection{Relativistic Systems}

For the case of a relativistic particle moving in a background spacetime,
we will enforce a constraint that effectively makes the $00$ component
of the metric tensor negative, and the $ij$ part ($i, j=1, 2, 3$) to 
consist of a positive semidefinite matrix. 
The relativistic Lagrangian as a function of the proper time 
is a Lorentzian length squared of the 4-velocity. For any given spacetime
point $p$, it is possible to choose a coordinate system (a Gaussian system)
that makes the components of the metric tensor equal to the Minkowski metric, 
$\operatorname{diag}(-1, 1, 1, 1)$. In a small enough neighborhood of $p$,
the $00$ component of the metric will be negative, and the $ij$ part
will consist of a positive-definite matrix, i.e., $[g_{ij}]\succ 0$. Thus, for any point of a spacetime, the metric tensor written in Gaussian coordinates will
satisfy the constraint $g_{00}<0$, $[g_{ij}]\succ 0$. In fact, 
such constraints are satisfied by a much wider class of settings than
small neighborhoods of Gaussian coordinates, most of the 
commonly encountered metric tensors in general relativity
satisfy these constraints (except when written in coordinate systems that include null coordinates).

With this motivation, we use a ReLU regularization that penalizes 
positive values for the $M_{00}$ component of the mass matrix, and 
deal with the spatial submatrix $M_{ij}$ ($i, j=1, 2, 3$) exactly as 
in the nonrelativistic case:

\begin{equation}
\begin{split}
\mathcal{L}_{\text{relativistic}} = \text{MAE}(\dots) &+ \lambda \left( \frac{\text{ReLU}(M_{00})}{c_0 + \epsilon} \right. \\
& \left. + \sum_{k=1}^{n-1} \frac{\text{ReLU}(-\det(M^{\text{spatial}}_k))}{s^{\text{spatial}}_k + \epsilon} \right)
\end{split}
\label{eq:sylvester_relativistic_loss}
\end{equation}
The temporal part is normalized by $c_0$, while spatial terms use $s^{\text{spatial}}_k = \prod_{i=1}^{k} c_i^{\text{spatial}}$.

This approach robustly enforces physical constraints across different mechanical regimes. While this method deals with the relativistic case rather 
naturally, it incurs $O(n^4)$ computational complexity, which is asymptotically higher than both eigenvalue decomposition and Hessian inversion ($O(n^3)$). However, this increased cost remains negligible for the small input dimensions typical in physics applications. 
Finally, note that this setting assumes knowledge of which coordinate
among the four coordinates is the timelike one.

\subsection{Gradient Clipping}

In LNNs, as derived in Appendix \ref{Appendix:A}, the gradient of the loss function contains the inverse Hessian term, making learning updates proportional to this term. 
When the mass matrix approaches singularity, weight and bias updates become arbitrarily large, typically resulting in exploding gradients.

To address this issue, we employ gradient clipping for preventing training destabilization. We clip the gradient norm at a threshold of 1.

\subsection{Physical Scaling}

While input normalization typically stabilizes neural network training and mitigates exploding or vanishing gradients, the physical context of LNNs requires special consideration. The inputs to LNNs -generalized coordinates and velocities- possess inherent physical units. Moreover, we compute accelerations according to the derivatives of the Lagrangian with respect to these coordinates and velocities.

We therefore implement a physics-aware normalization scheme, effectively equivalent to unit conversion while preserving temporal units (or proper time units for relativistic inputs). For each dimension (i.e., $q_i$ and $\dot{q}_i$), we identify the maximum absolute value in the training dataset and use these values as scaling factors. We apply this normalization to both training and test datasets:
\begin{equation}
\tilde{q}_i = \frac{q_i}{\max_{t \in \mathcal{T}_{\text{train}}} |q_i(t)|}, \quad \tilde{\dot{q}}_i = \frac{\dot{q}_i}{\max_{t \in \mathcal{T}_{\text{train}}} |\dot{q}_i(t)|}
\end{equation}
We maintain these scaling factors to transform predicted accelerations back to their original units after inference.

\subsection{Activation Functions} \label{subsec:activationmethods}

Activation functions constitute one of the most critical architectural decisions. While \citet{cranmerLagrangianNeuralNetworks2020} selected Softplus as their activation function, \citet{Liu:2021xwt} argued that a hybrid network, with half the neurons using quadratic functions and half using Softplus, yields superior performance.
We contend that neither approach is optimal for LNN architectures and identify GeLU as the most effective activation function.

Additionally, we propose a novel activation function as an alternative to the quadratic function:
\begin{equation}
f(x) = x \tanh(kx), \quad k = 0.5
\end{equation}
This function initially behaves quadratically near zero but transitions to linear behavior for larger inputs. While it significantly outperforms Softplus, quadratic, and hybrid variants, it remains slightly inferior to GeLU in our experiments.

\subsection{Network Initialization}

Network initialization critically influences model stability, as suboptimal initialization can trigger exploding or vanishing gradients. Standard initialization schemes in machine learning libraries aim to preserve variance across layers during either forward or backward passes, essentially managing gradient statistics. The optimal initialization strategy depends on the chosen activation function; for instance, Kaiming initializations are tailored for ReLU-like activations.

LNNs present unique challenges with their unconventional forward and backward passes. This motivated \citet{cranmerLagrangianNeuralNetworks2020} to develop custom initialization schemes, while \citet{Liu:2021xwt} retained default initialization but ensured network outputs begin near zero, then adding an initial quadratic velocity term to guarantee an invertible, positive semi-definite mass matrix at the start of the training.

We experimented with an orthodox approach by implementing variance-preserving initialization for the backward pass, tailored for our activation functions. Our approach can be thought of as generalizing the work of \citet{He:2015dtg} to other activation functions, which is identical to what \citet{binghamAutoInitAnalyticSignalPreserving2022} does. The custom initialization benefits certain activation functions like our proposed $x\tanh(kx)$, though it slightly degrades GeLU performance.

We also incorporate the quadratic velocity term, as our experiments demonstrate improved stability across various activation functions and initialization schemes.

\subsection{Training Range}\label{subsec:training_bounds}

Our experiments reveal that training networks on the exact testing range can degrade the Hessian behavior near boundary values. We therefore employ extended training ranges in scenarios where this does not compromise black-box Lagrangian discovery. For angular coordinates, instead of training on $[0, 2\pi]$, we observe improvements when training on $[-\epsilon, 2\pi + \epsilon]$ where $\epsilon$ is between $\pi/10$ and $\pi/2$. Since we assume knowledge of angular periodicity of our inputs, this extended range maintains methodological integrity.

\textbf{Implementation Note:} When solving trajectories, we apply modulo $2\pi$ to angular coordinates before model input to prevent catastrophic out-of-distribution errors. This proves essential as standard ODE solvers allow angles to grow unboundedly.

\begin{figure*}[htbp]
    \centering
    \includegraphics[width=\textwidth]{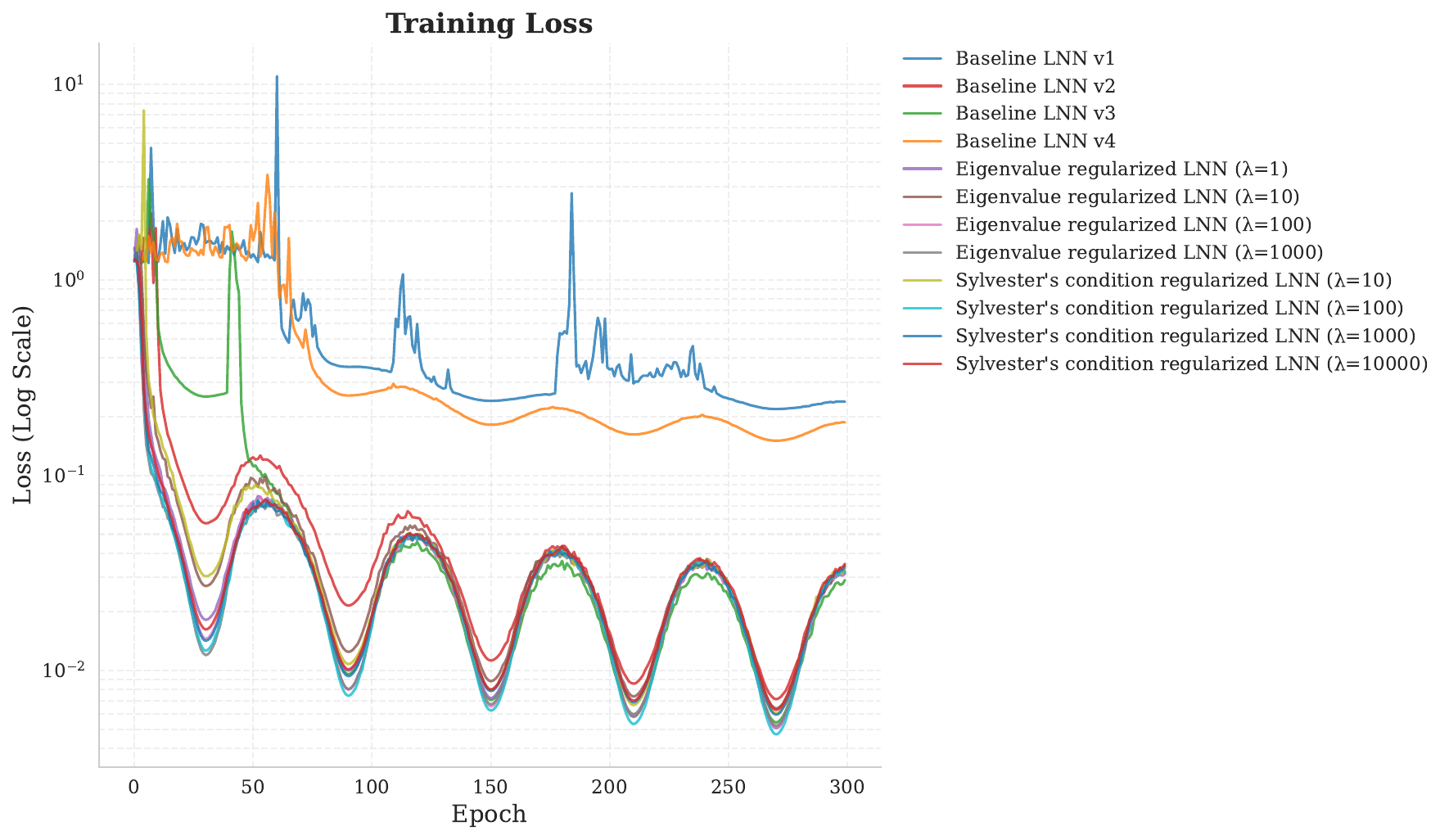}
    \caption{\capRegularizationExistence}
    \label{fig:regularization_existence}
\end{figure*}

\section{Network Experiments} \label{sec:networkexp}

We tested how our proposed changes affect the ability to successfully represent the underlying physical systems. We detail our training specifications in Appendix~\ref{Appendix:TrainingDetailsNetwork}, and for our definition of instability, see Appendix~\ref{Appendix:Instability}.

\subsection{Effect of the Hessian Regularization Term.}

To evaluate the Hessian regularization methods introduced in Section \ref{subsec:hessian}, we conducted experiments on a double pendulum system. We trained the network configurations \trainingTimesX{} times using: only the standard MAE loss, eigenvalue regularization, Sylvester's condition regularization, varying the values of $\lambda$.

Our results demonstrate significant improvements:

The LNN with eigenvalue regularization achieved validation loss values \percentageOne\% lower than those of the standard unregularized LNN, while the Sylvester regularization approach yielded \percentageTwo\% lower loss values compared to the baseline, on average, as seen in Figure~\ref{fig:regularization_existence}. While the unregularized model failed to learn sufficiently for successful ODE solver integration in 2 out of 4 tests, both regularization methods achieved performance well above adequate levels. Average training instability decreased by \percentageThree\% and \percentageFour\% for eigenvalue and Sylvester regularization, respectively, relative to the unregularized model, which is seen in Figure~\ref{fig:regularization_comparison}.

Although the presence of regularization significantly enhances both training stability and overall performance, the performance difference between our two proposed regularization techniques is negligible, as seen in Figure~\ref{fig:lambda_values}. Optimal performance was achieved with $\lambda = \lambdaEigenOptimal$ for eigenvalue regularization and $\lambda = \lambdaSylvesterOptimal$ for Sylvester regularization on the double pendulum system. Our experiments reveal that regularization hyperparameters can be set above these optimal values with minimal performance degradation. The ideal network behavior involves rapidly driving the regularization term to zero, effectively continuing with the MAE loss thereafter. Given the varying complexity across systems, we recommend initiating training with higher $\lambda$ values to ensure the early learning of the correct mass matrix signature.

\subsection{Effect of Scaling.}

We conducted double pendulum experiments to evaluate how our proposed physical scaling system affects performance. We tested the LNN without any scaling, the LNN with physical scaling, and the LNN with DyT, which was recently proposed as an alternative to BatchNorm/LayerNorm in large language models \citep{250310622TransformersNormalization}.

By definition, the scaling factor for each coordinate in our method is inversely proportional to the maximum value of the inputs. In double pendulum systems, all coordinates share the same units (angles) and scaling value (10, derived from the max\_velocity chosen for both coordinates). This allows us to scale the MAE loss and training instability terms and directly compare results of the experiments using what we term the normalized loss and normalized instability.

\begin{figure}[htbp]
\centering
\includegraphics[width=0.48\textwidth]{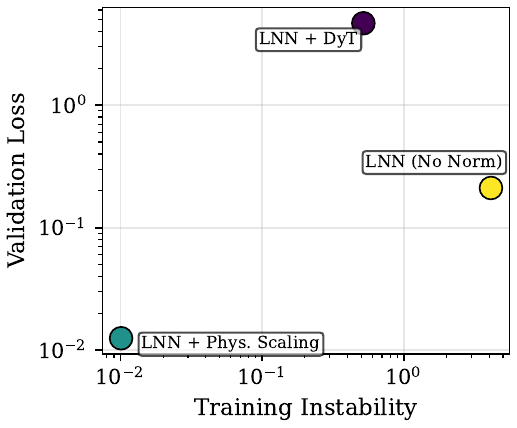}
\caption{\capScalingPerformance}
\label{fig:scaling_performance}
\end{figure}

Our results show that the normalized loss of the physically scaled system is \percentageFive\% lower than those of the unscaled version. The instability score of the scaled system also demonstrates significant improvement (\percentageSix\% lower) compared to the unscaled version. DyT did not improve performance in our experiments and fell far short of serving as a replacement for our proposed physical scaling method. (See Figure~\ref{fig:scaling_performance})

\subsection{Effect of the Initial Quadratic Term.}

We conducted experiments to see whether adding an initial quadratic term as suggested by \cite{Liu:2021xwt} will improve performance. We tested various network configurations with and without the initial quadratic term.

\begin{figure}[htbp]
\centering
\includegraphics[width=0.48\textwidth]{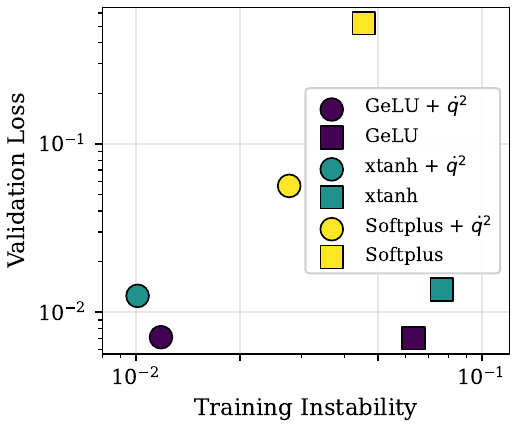}
\caption{\capExtraTerm}
\label{fig:extra_term_performance}
\end{figure}

As seen in Figure~\ref{fig:extra_term_performance}, including an initial quadratic term substantially lowers the training instability, as well as the average validation losses.

\subsection{Activation Function Experiments.}

We conducted experiments with the following activation functions: Softplus, Softplus + $x^2$ (Quadratic Activation), GeLU, and the proposed $x\tanh(kx)$ introduced in \ref{subsec:activationmethods}.

Our results demonstrate that GeLU achieved the best overall performance with \avgLossGeLU{} average validation loss, with $x\tanh(kx)$ following closely behind with \avgLossXTanhKX{}. The Softplus and Softplus + Quadratic hybrid networks exhibited substantially lower performance (\avgLossSoftplus{} and \avgLossHybrid{} average validation loss respectively) and suffered from relatively worse training instabilities, as shown in Table~\ref{tab:activation_comparison}, and evident from the training loss over epoch graphs (See Figure~\ref{fig:activation_comparison}). These findings underscore the critical importance of activation function selection for training stability.

\begin{table}[t]
\centering
\begin{tabular}{lcc}
\toprule
\textbf{Experiment} &
\makecell{\textbf{Average} \\ \textbf{Validation Loss}} &
\makecell{\textbf{Average} \\ \textbf{Instability}} \\
\midrule
GeLU & \textbf{0.00738} & \textbf{0.01452} \\
$x\tanh(kx)$ & 0.01114 & 0.05048 \\
Softplus & 0.04452 & 0.02850 \\
Softplus + Quadratic Hybrid & 0.16380 & 0.04915 \\
\bottomrule
\end{tabular}

\caption{\capActivationTable}
\label{tab:activation_comparison}
\end{table}

\subsection{Weight Initialization Experiments.}

We conducted experiments using two approaches: (1) without any custom initialization, and (2) with gradient variance preserving initialization tailored for each activation function.
Our custom initialization improved performance for Softplus, Softplus-Quadratic hybrid model and $x\tanh(kx)$, but slightly reduced performance for GeLU, as shown in Table~\ref{tab:initialization_comparison}.

This result may be partially explained by the fact that default initializations, such as the Kaiming initialization \citep{He:2015dtg}, are optimized for ReLU and ReLU-like activation functions, such as GeLU. 

\begin{table}[htbp]
\centering
\begin{tabular}{lcc}
\toprule
\textbf{Experiment} & \textbf{Val. Loss} & \textbf{Train Instability} \\
\midrule
GeLU v2 + $\dot{q}^2$ & \textbf{0.006859} & 0.02867 \\
GeLU + $\dot{q}^2$ & 0.007114 & 0.01179 \\
\makecell[l]{GeLU with custom\\initialization + $\dot{q}^2$} & 0.007649 & 0.01726 \\
\makecell[l]{GeLU v2 with custom\\initialization + $\dot{q}^2$} & 0.009203 & \textbf{0.006530} \\
\makecell[l]{$x\tanh(kx)$ with custom\\initialization + $\dot{q}^2$} & 0.009808 & 0.09090 \\
$x\tanh(kx)$ + $\dot{q}^2$ & 0.012475 & 0.01007 \\
\makecell[l]{Softplus with custom\\initialization + $\dot{q}^2$} & 0.032768 & 0.02932 \\
\makecell[l]{Quadratic-Softplus Hybrid\\with custom init. + $\dot{q}^2$} & 0.050726 & 0.03244 \\
Softplus + $\dot{q}^2$ & 0.056272 & 0.02768 \\
Quadratic-Softplus\\ Hybrid + $\dot{q}^2$ & 0.276867 & 0.06585 \\
\bottomrule
\end{tabular}
\caption{\capInitializationTable}
\label{tab:initialization_comparison}
\end{table}

\begin{figure*}[htbp]
\centering
\includegraphics[width=\textwidth]{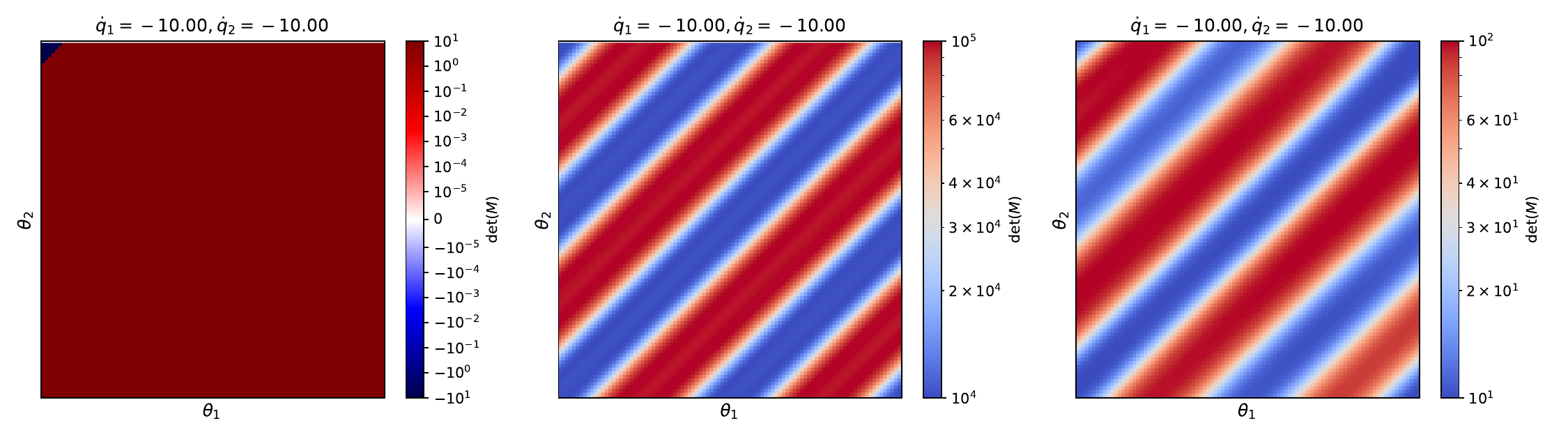}
\caption{\capTrainingRange}
\label{fig:training_range_comparison}
\end{figure*}

\subsection{Training Range Experiments.}

To help the network learn the periodicity of periodic inputs, we experimented with different bounds for the angle coordinates in the double pendulum system training dataset. Training on $[0, 2\pi]$ can cause the learned Hessian (mass matrix) to exhibit problematic behavior, such as incorrect determinant signs near $0$ and $2\pi$, particularly when velocities approach their training bounds. Expanding the training range to $[-\epsilon, 2\pi + \epsilon]$, where $\epsilon$ is between $\pi/10$ and $\pi/2$, eliminates this issue, as seen in Figure~\ref{fig:training_range_comparison}.

\subsection{Comparison with the Previously Suggested Tricks.}

We have compared our best model with a model that we trained using the initialization and the activation function proposed by \cite{cranmerLagrangianNeuralNetworks2020}, as well as a model that we trained following the tricks suggested by \cite{Liu:2021xwt}.

\begin{figure}[htbp]
\centering
\includegraphics[width=0.48\textwidth]{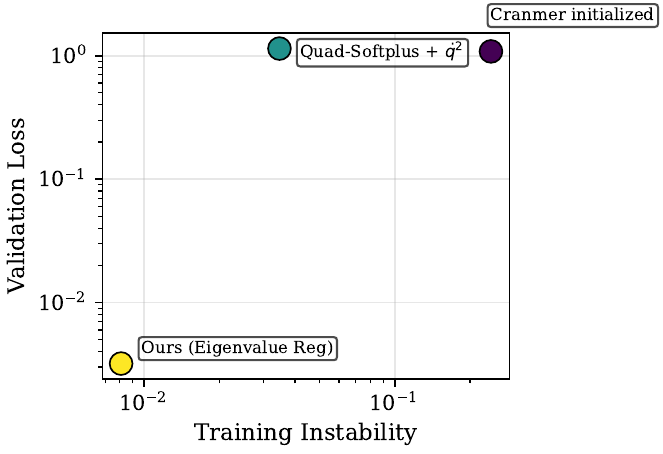}
\caption{\capOldModels}
\label{fig:old_models}
\end{figure}

As demonstrated in Figure~\ref{fig:old_models}, the results clearly show significant improvements on both stability of the model, as well as the validation losses.

\FloatBarrier

\section{Physical Experiments} \label{sec:physicalexp}

\begin{figure*}[htbp]
\centering
\includegraphics[width=\textwidth]{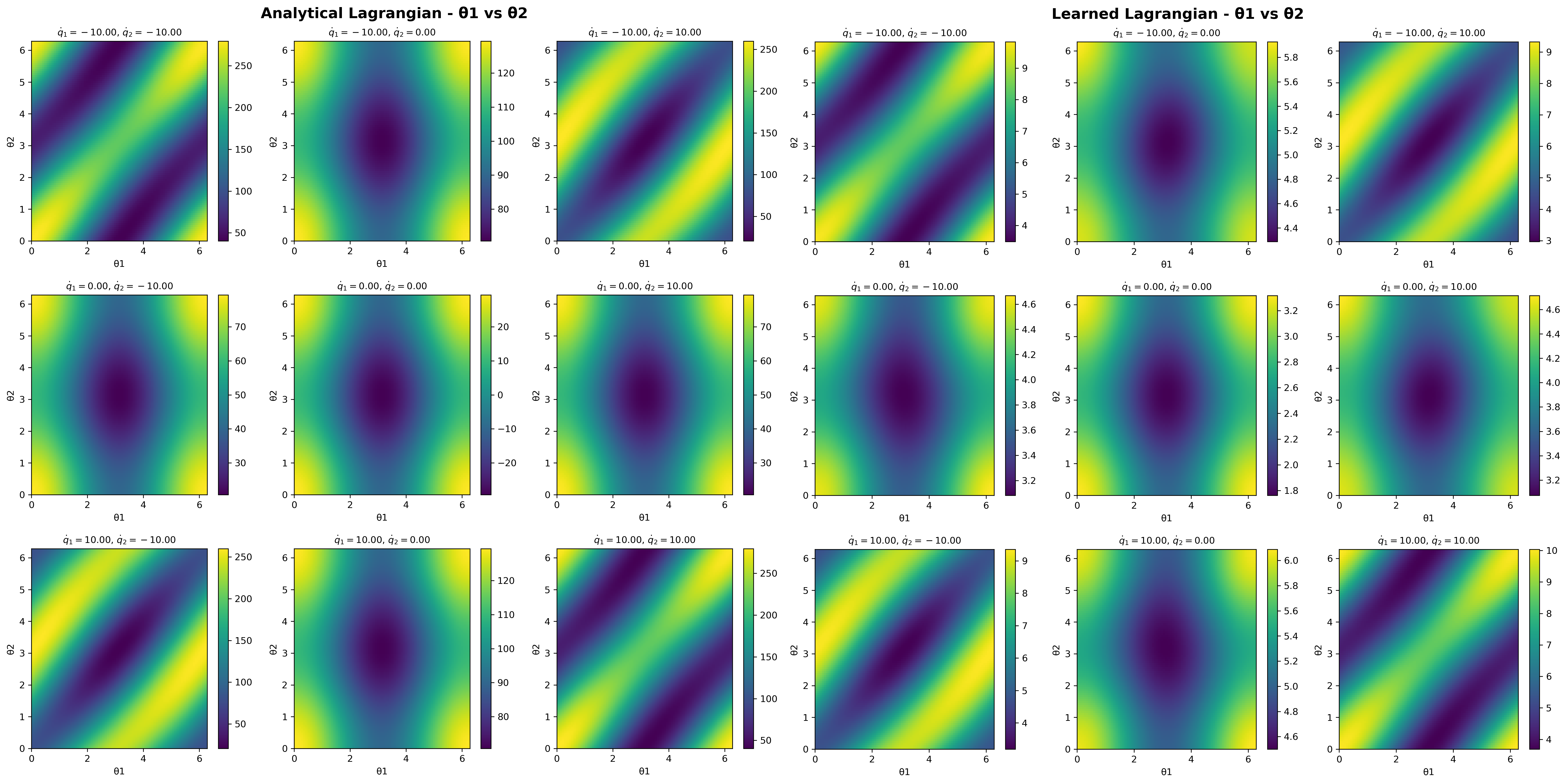}
\caption{\capDoublePendulumLagrangian}
\label{fig:double_pendulum_lagrangian}
\end{figure*}

We tested our proposed network in a variety of physical systems to showcase its ability to learn complex behaviour. We tested our network against a general relativistic system, which is a new contribution to the literature due to the appearance of time as a component in the generalized coordinates.

The Lagrangians and other training details for the systems are available at Appendix~\ref{Appendix:TrainingDetailsPhysical}.

\subsection{Double Pendulum}

The double pendulum system is the most complicated system that appeared in the original work. Our best model achieved \lOneLossOurs{} $L_1$ loss (scaled up to match the units with the original), a \percentageSeven\% improvement compared to their reported \lOneLossOriginal{} $L_1$ loss.

A given dynamical system such as the double pendulum can be described
by a set of different Lagrangians, even including examples
that aren't related to each other via simple scalings or the 
addition of total time derivatives. However, despite this freedom,
our results imply that our model learns the same Lagrangian (see Figure~\ref{fig:double_pendulum_lagrangian}) and the same 
Lagrangian Hessian (see Figure~\ref{fig:double_pendulum_hessian}) as the analytical solution, up to a simple scaling.

\subsection{Spring Pendulum}

The spring pendulum system is a challenging system to learn that consists of a pendulum whose length obeys Hooke's law. The acceleration of the system can be written analytically as:
\begin{align}
\ddot{r} &= r\dot{\theta}^2 + g\cos(\theta) - \frac{k}{m}(r-1) \\
\ddot{\theta} &= \frac{-g\sin(\theta) - 2\dot{r}\dot{\theta}}{r}
\end{align}

The system has a singularity at $r = 0$. Our experiments show that the proposed LNN model can learn the system very well if we avoid approaching the singularity point. With the regularization term added, even though there were spikes, the training proceeded successfully. We ran the system with 4 different random configurations, and the average energy error is \energyErrorPercentage\% of the energy of the system after after $t = 50s$.

\begin{figure}[htbp]
\centering
\includegraphics[width=0.48\textwidth]{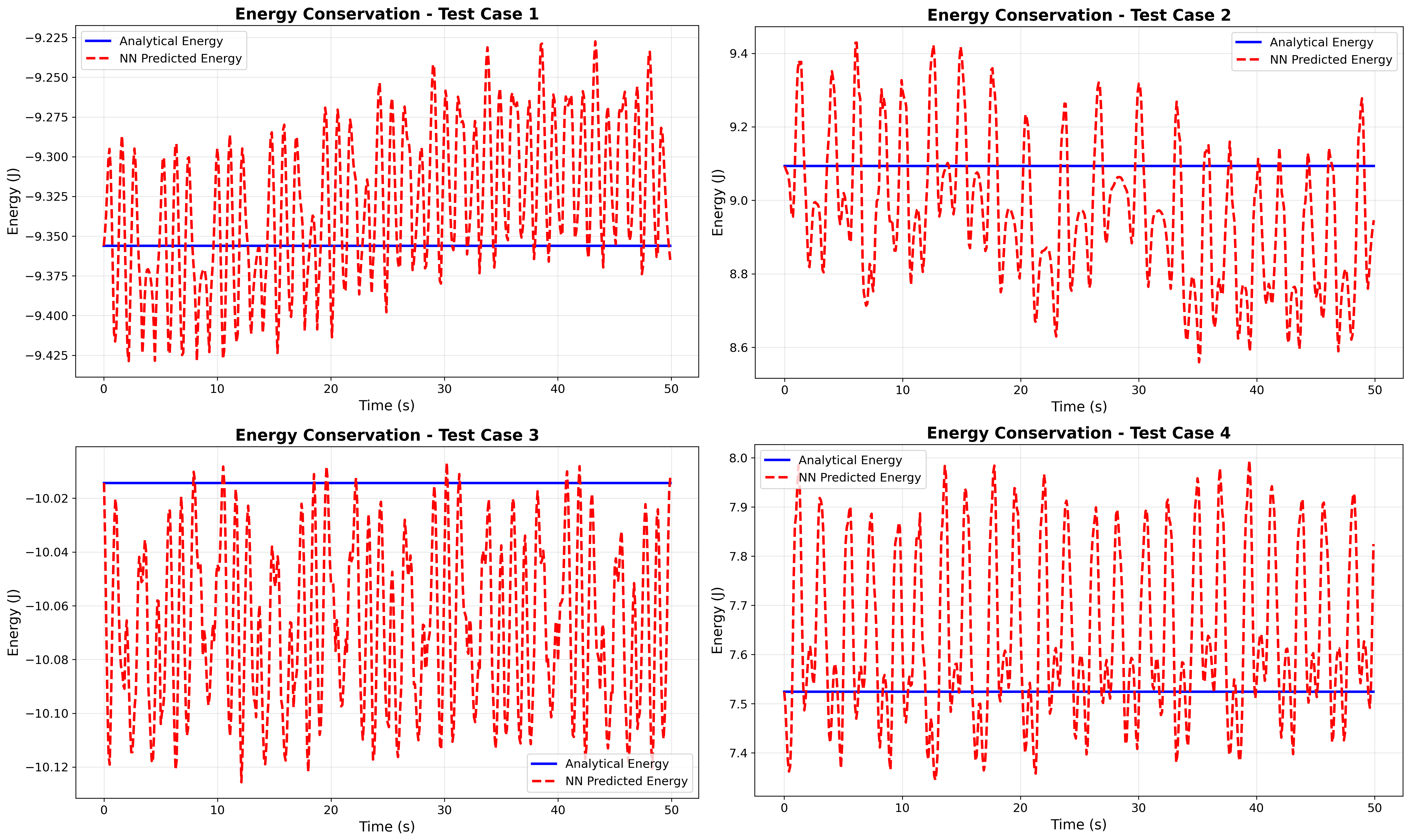}
\caption{\capSpringPendulumEnergyError}
\label{fig:spring_pendulum_energy_error}
\end{figure}

\begin{figure}[htbp]
\centering
\includegraphics[width=0.48\textwidth]{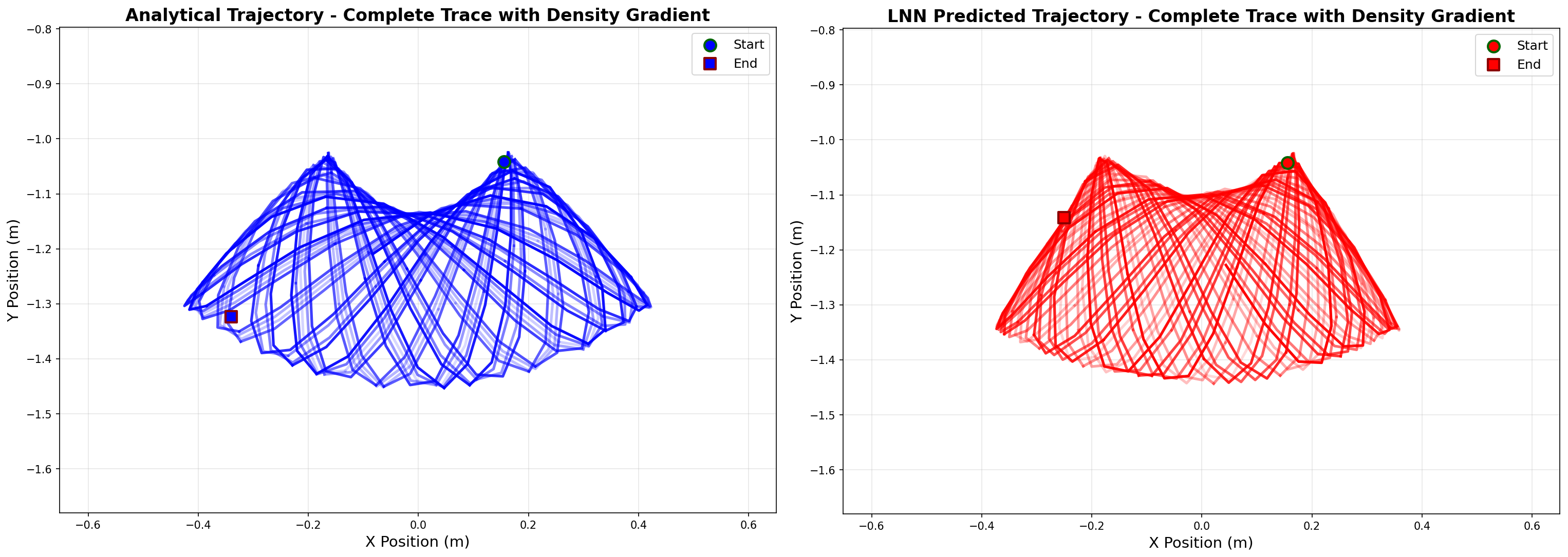}
\caption{\capSpringPendulumTrace}
\label{fig:spring_pendulum_trace}
\end{figure}

\subsection{Triple Pendulum}

The triple pendulum presents an even more challenging test for our improved network. The only instance we found of a triple pendulum tested with an LNN-based network was the CLNN model \citep{finziSimplifyingHamiltonianLagrangian2020}, which employs significant architectural changes, such as the the requirement of having cartesian coordinates and knowledge of the holonomic constraints beforehand.

\begin{figure}[htbp]
\centering
\includegraphics[width=0.48\textwidth]{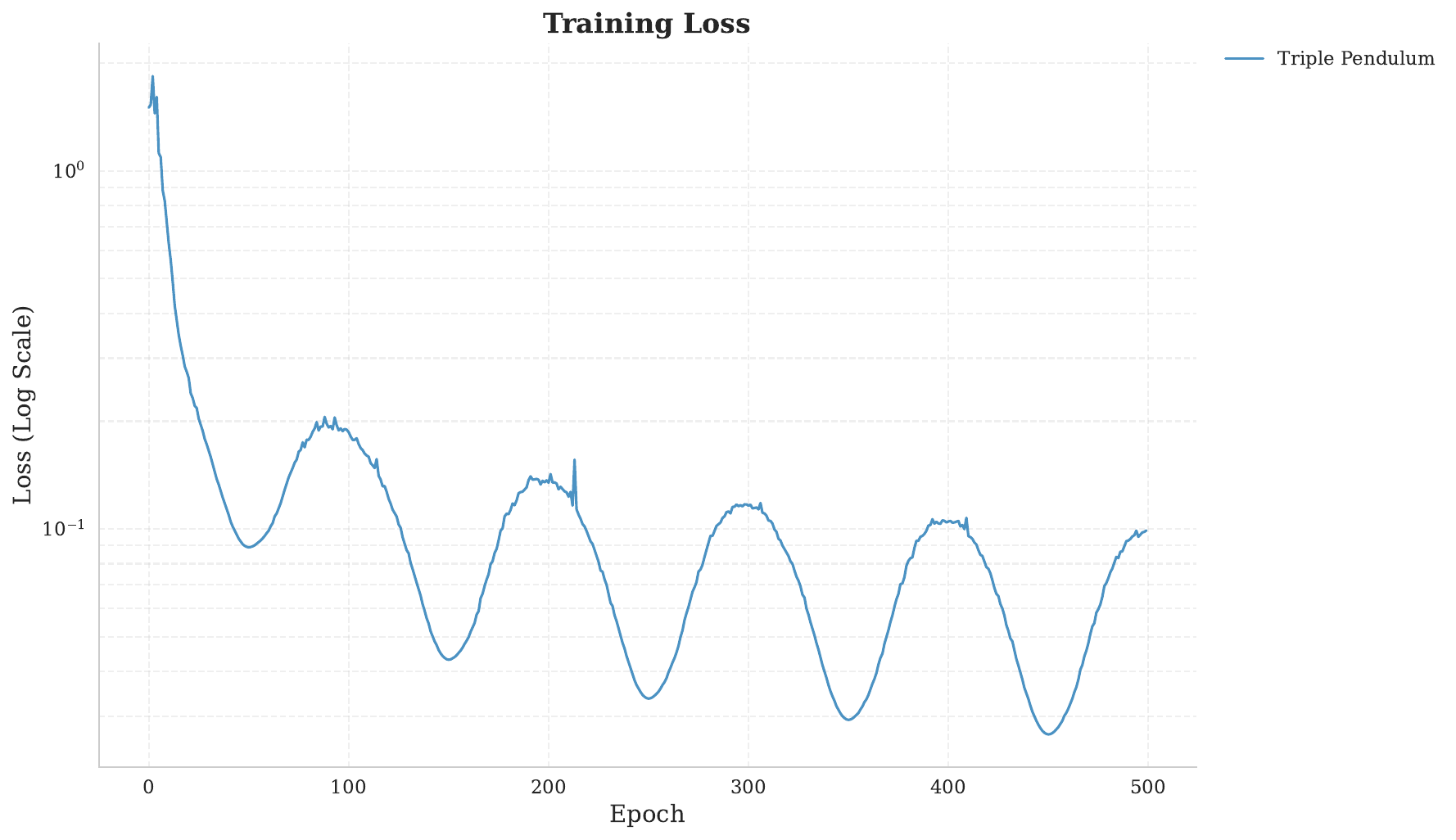}
\caption{\capTriplePendulumTraining}
\label{fig:triple_pendulum_training}
\end{figure}

\begin{figure}[htbp]
\centering
\includegraphics[width=0.48\textwidth]{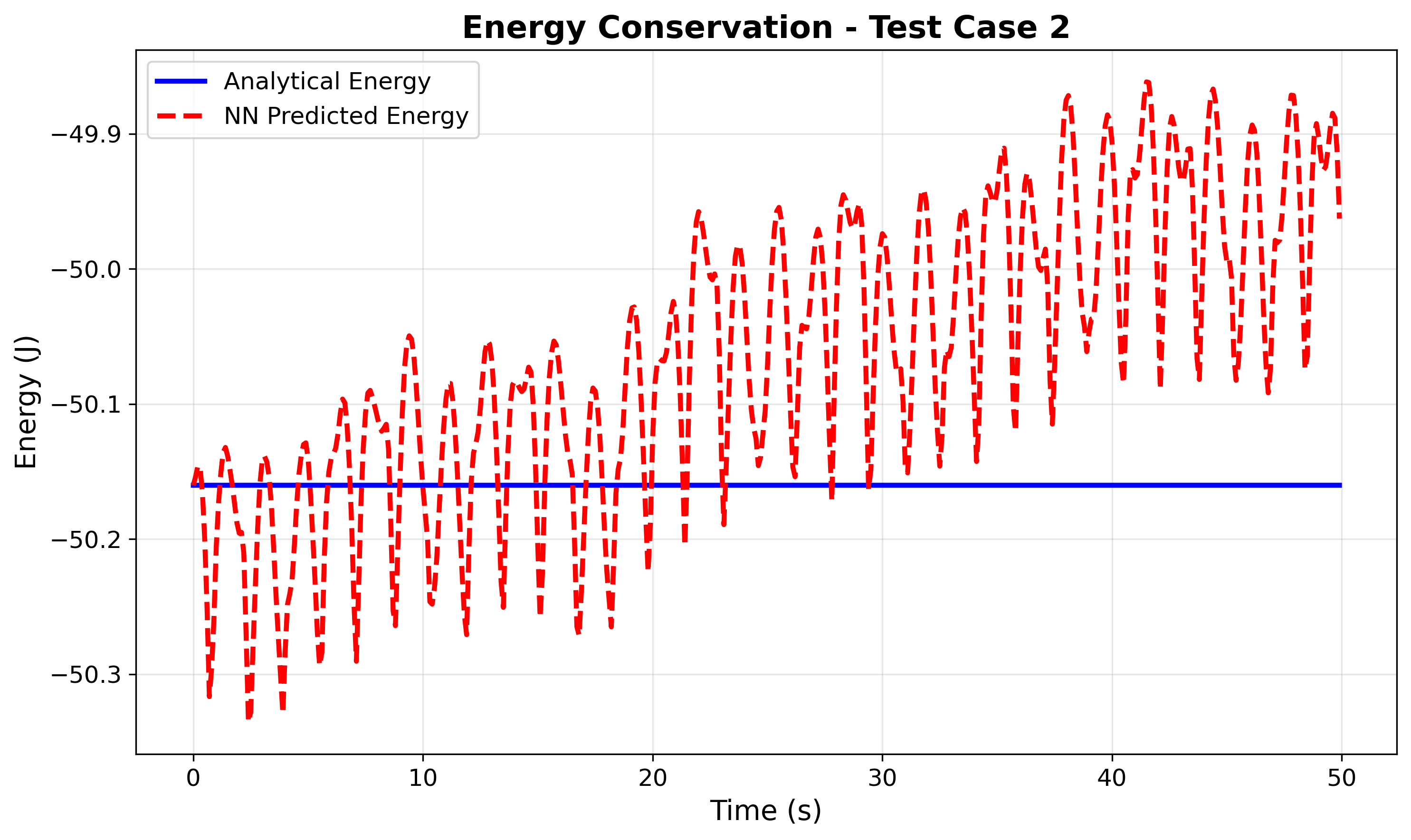}
\caption{\capTriplePendulumEnergyNonChaotic}
\label{fig:triple_pendulum_energy_nonchaotic}
\end{figure}

Our training was very smooth and the model converged relatively quickly, starting from \validationLossTripleStart{} validation $L_1$ loss in the first epoch and converging up to \validationLossTriple{} in the end. The training process was remarkably smooth (see Figure~\ref{fig:triple_pendulum_training}). However, 3 out of 4 randomly initialized tests entered the chaotic regime. Since the system exhibits highly chaotic behavior, predictions diverge relatively quickly. This is not a network limitation since we observe the same behavior across different analytical ODE solver instances. In the non-chaotic regime test, predictions remained stable until the very end of the test, with an energy error of \energyErrorNonChaotic\%, as seen in Figure~\ref{fig:triple_pendulum_energy_nonchaotic}.

\FloatBarrier

\subsection{Geodesic on a Sphere}

Since the network learns a general Lagrangian function, we can use it to predict geodesic paths in both classical and relativistic settings, as geodesics in semi-Riemannian manifolds are always expressible with a kinetic term as the Lagrangian \citep{oneillSemiRiemannianGeometryApplications1983}. 

To test this, we created geodesic trajectory data for a particle embedded on the surface of a sphere. Its Lagrangian is:
\begin{align}L = \frac{1}{2} m R^2 \left(\dot{\theta}^2 + \sin^2\theta \, \dot{\phi}^2\right)
\end{align}

The inputs are $(\theta, \phi, \dot{\theta}, \dot{\phi})$. The system is challenging to train because of the $\theta = 0$ coordinate singularity. The geodesic equations on the sphere surface are
\begin{align}
\ddot{\theta} &= \sin\theta \, \cos\theta \, \dot{\phi}^2, \\
\ddot{\phi} &= -2 \, \frac{\cos\theta}{\sin\theta} \, \dot{\theta} \, \dot{\phi}.
\end{align}

Here, $\ddot{\theta}$ and $\ddot{\phi}$ represent the accelerations along the $\theta$ and $\phi$ directions, respectively. $\ddot{\phi}$ is undefined when $\theta = 0$ and becomes arbitrarily large as $\theta$ approaches 0. Therefore, we must maintain a safe distance from the singularity point during training and testing.

The network was able to learn the system dynamics relatively quickly, 
with negligible error in both energy and position.

\begin{figure}[htbp]
\centering
\includegraphics[width=0.48\textwidth]{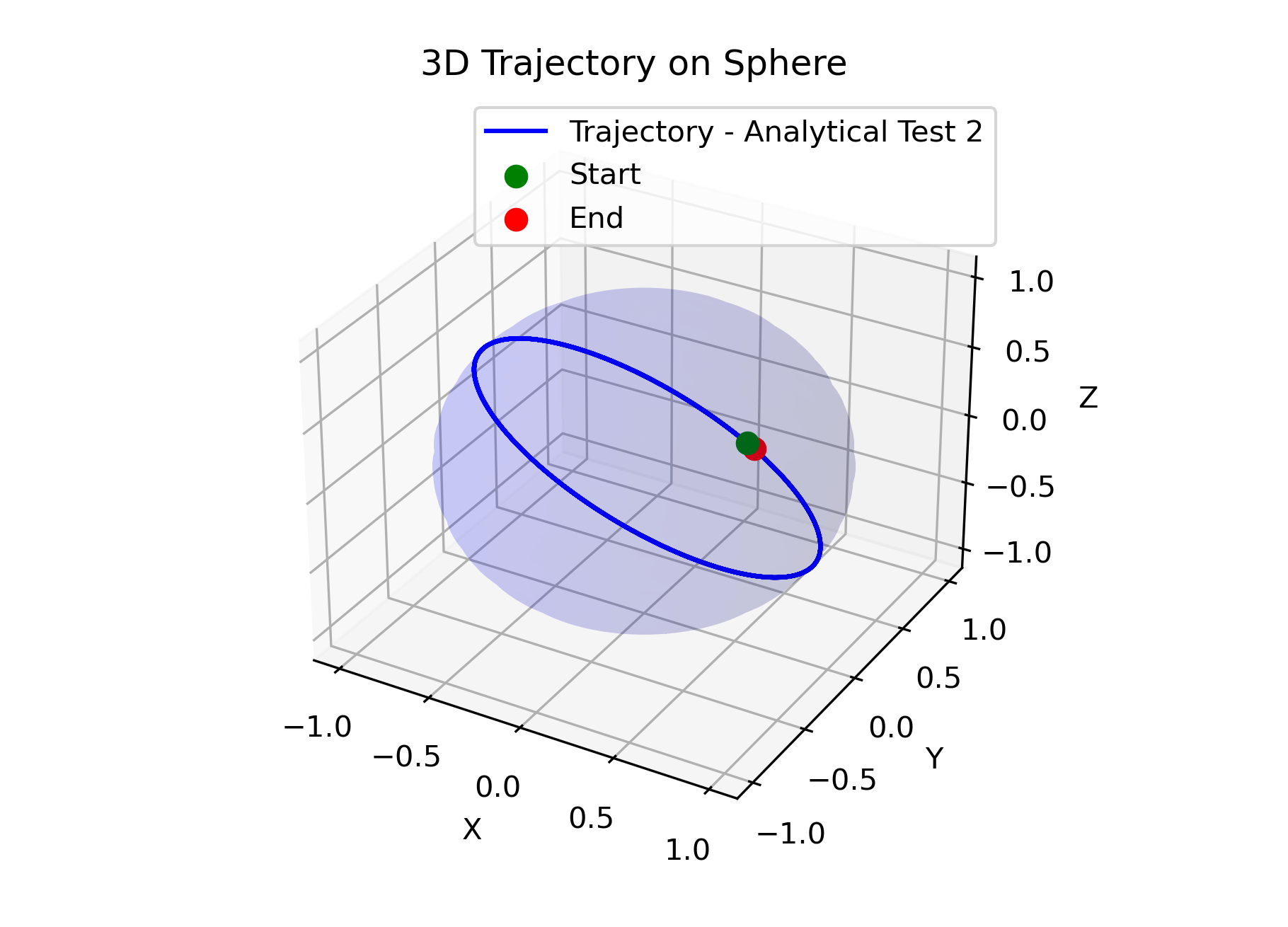}
\caption{\capSphereCoordAnalytical}
\label{fig:sphere_coord_analytical}
\end{figure}

\begin{figure}[htbp]
\centering
\includegraphics[width=0.48\textwidth]{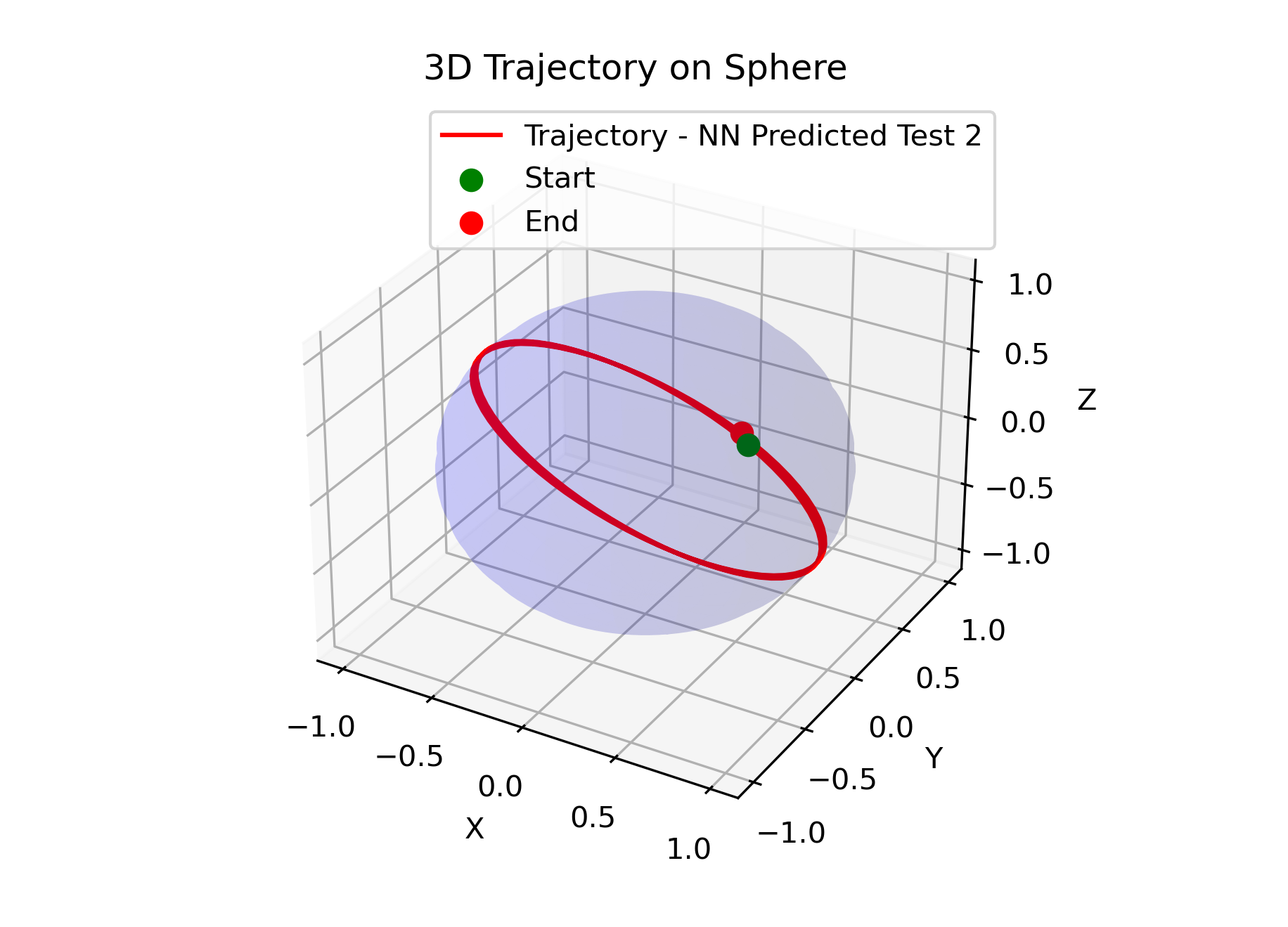}
\caption{\capSphereCoordLearned}
\label{fig:sphere_coord_learned}
\end{figure}

\subsection{Relativistic: AdS\textsubscript{4}}
\begin{figure}[htbp]
\centering
\includegraphics[width=0.48\textwidth]{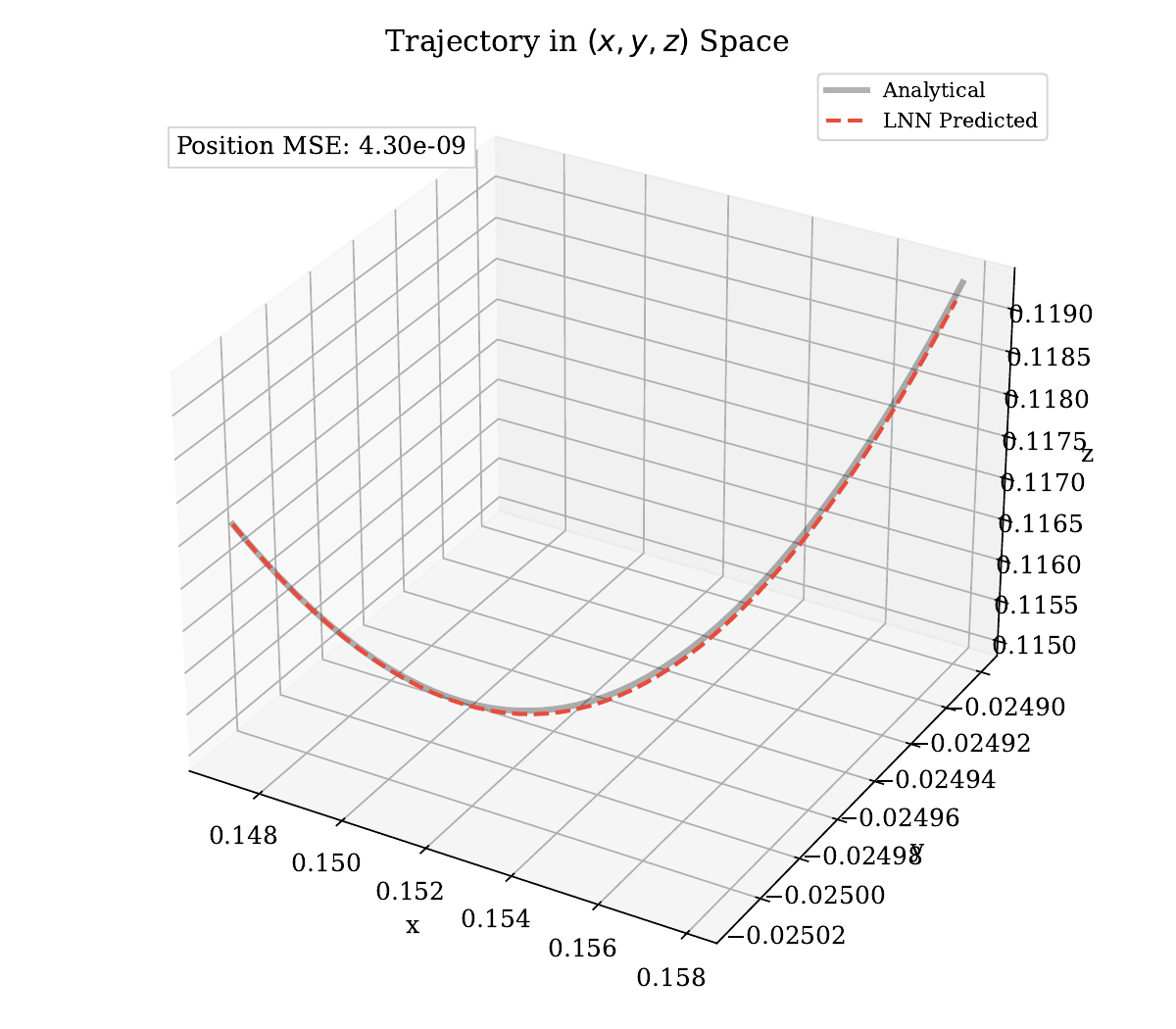}
\caption{\capAdSCoordError}
\label{fig:ads4_coord_error}
\end{figure}

\begin{figure}[htbp]
\centering
\includegraphics[width=0.48\textwidth]{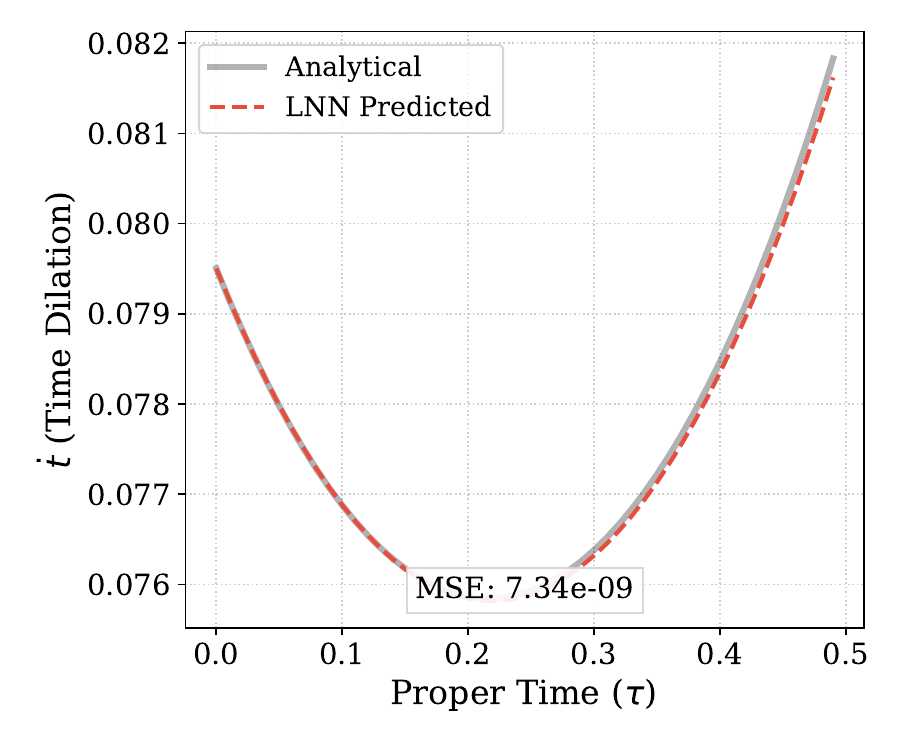}
\caption{\capAdSTDot}
\label{fig:ads4_tdot}
\end{figure}

To showcase our regularization method that enforces Lorentzian signature for relativistic systems, dubbed ``Lorentzian Sylvester regularization,'' we tested particle geodesics on the 4-dimensional anti-de Sitter space, 
i.e., AdS4, using the ``half-space'' coordinates analogous to the
Poincar\`e half-space model of the hyperbolic space. In these coordinates,
the metric tensor is given as,
\begin{align}
    ds^2 = \frac{1}{z^2} (-dt^2 + dx^2 + dy^2 + dz^2)\,.
\end{align}
In this case, the input $\mathbf{q}, \dot{\mathbf{q}}$ consists of $(t, x, y, z, \dot{t}, \dot{x}, \dot{y}, \dot{z})$, where time derivatives are taken with respect to proper time $\tau$. In the testing phase, all short-time trajectory tests conducted closely followed the actual trajectories (see Figure~\ref{fig:ads4_coord_error}), as well as the time dilation factor $\dot{t}$ (see Figure~\ref{fig:ads4_tdot}) and $\dot{z}$ (see Figure~\ref{fig:ads4_zdot})

\section{Conclusion \& Discussion} \label{sec:conclusion}

We have proposed a novel regularization scheme for Lagrangian Neural Networks that accommodates both classical and relativistic systems by enforcing appropriate mass matrix signatures. We have also introduced several improvements, including GeLU activation functions and physical scaling, which resulted in significantly improved stability and enabled us to train on more complex systems such as triple pendulums. Additionally, we systematically evaluated previous stability techniques proposed in the literature. More importantly, we demonstrated that LNNs with space-time coordinate inputs can learn geodesic trajectories and extract metric tensors from scalar Lagrangians, applications we have not encountered in previous literature.

This approach enables discovery of metric tensor elements through the learned mass matrix. For systems where the dimensionality is sufficiently low, symbolic regression can reveal the analytical forms.

It should be noted that certain complex systems were hard to train, such as for the particle following a geodesic in the Schwarzschild metric \citep{schwarzschildUberGravitationsfeldMassenpunktes1916}, especially when the particle is close to the event horizon.

Our proposed model also inherits certain limitations from the original LNN architecture, notably the inability to accommodate constraints and the requirement for invertible Hessians throughout training. Certain coordinate choices, including redundant coordinates or those with singularities as in our sphere geodesic example, can produce physically valid but non-invertible Hessians, for which the current model remains unsuitable. 

While CLNNs \citep{finziSimplifyingHamiltonianLagrangian2020} deal with the issues arising from the choice of coordinates through explicit holonomic constraints in cartesian coordinates, this approach requires prior system knowledge that may be available in robotics applications but contradicts the black-box paradigm essential for physical law discovery. Therefore, a natural extension would be developing LNN architectures capable of automatically discovering and incorporating constraints while preserving the black-box learning approach.

\FloatBarrier

\bibliographystyle{apsrev4-2}
\bibliography{references}

\appendix

\section{Derivation of Gradient of Loss for the Lagrangian Neural Network}\label{Appendix:A}

This appendix provides a detailed derivation of the gradient of the total loss function with respect to an arbitrary network parameter $\theta$ for Lagrangian Neural Network architecture.

\subsection*{A.1 Model Formulation}

The LNN model first uses a Multi-Layer Perceptron (MLP), parameterized by a set of weights and biases $\theta$, to approximate the scalar Lagrangian $\mathcal{L}(\mathbf{q}, \dot{\mathbf{q}}; \theta)$.
The predicted acceleration $\ddot{\mathbf{q}}_{\text{predicted}}$ is then computed by solving the Euler-Lagrange equation, given in \eqref{eq:acceleration}.

The network is trained by minimizing a loss function that compares the predicted acceleration to the true acceleration, plus the Hessian regularization \eqref{eq:eigen_loss}.
For this derivation, we only look into the Mean Absolute Error (MAE) part of the loss:
\begin{align}
\mathcal{L}_{\text{MAE}} &= \text{MAE}(\ddot{\mathbf{q}}_{\text{real}}, \ddot{\mathbf{q}}_{\text{predicted}}) \nonumber \\
&= \frac{1}{N} \sum_{i=1}^{N} |\ddot{\mathbf{q}}_{\text{real}, i} - \ddot{\mathbf{q}}_{\text{predicted}, i}| \label{A.1}
\end{align}

Our goal is to derive the expression for $\frac{\partial \mathcal{L}_{\text{MAE}}}{\partial \theta}$, to see which terms the gradient of the loss depends on.

\subsection*{A.2 Simplified Notation}

To simplify the derivation, we introduce the following notation:
\begin{itemize}
    \item The predicted acceleration vector: $\mathbf{a} \equiv \ddot{\mathbf{q}}_{\text{predicted}}$
    \item The Hessian of the Lagrangian w.r.t. velocity (mass matrix): $\mathbf{H} \equiv \nabla_{\dot{\mathbf{q}}}\nabla_{\dot{\mathbf{q}}}^{\top}\mathcal{L}$
    \item The gradient of the Lagrangian w.r.t. position: $\mathbf{g} \equiv \nabla_{\mathbf{q}} \mathcal{L}$
    \item The mixed-derivative matrix associated with Coriolis and centrifugal forces: $\mathbf{C} \equiv \nabla_{\mathbf{q}}\nabla_{\dot{\mathbf{q}}}^{\top}\mathcal{L}$
\end{itemize}

Using this notation, Equation \eqref{eq:acceleration} becomes:
\begin{align}
\mathbf{a} = \mathbf{H}^{-1}[\mathbf{g} - \mathbf{C}\dot{\mathbf{q}}] \label{A.2}
\end{align}

Note that $\mathbf{H}$, $\mathbf{g}$, and $\mathbf{C}$ are all functions of the MLP output $\mathcal{L}$ and are therefore implicitly dependent on the parameter $\theta$.

\subsection*{A.3 Gradient Derivation}

We compute the gradient using the chain rule:

\begin{align}
\frac{\partial \mathcal{L}_{\text{MAE}}}{\partial \theta} = \frac{\partial \mathcal{L}_{\text{MAE}}}{\partial \mathbf{a}} \frac{\partial \mathbf{a}}{\partial \theta} \label{A.3}
\end{align}

First, we compute the gradient of the MAE loss $\mathcal{L}_{\text{MAE}}$ with respect to the predicted acceleration vector $\mathbf{a}$.
This yields the initial gradient vector for backpropagation.
\begin{align}
\frac{\partial \mathcal{L}_{\text{MAE}}}{\partial \mathbf{a}_i} = -\frac{1}{N} \text{sign}(\ddot{\mathbf{q}}_{\text{real}, i} - \mathbf{a}_i) \label{A.4}
\end{align}

For clarity, we denote this upstream gradient vector as $\boldsymbol{\delta}_{\mathbf{a}} \equiv \nabla_{\mathbf{a}} \mathcal{L}_{\text{MAE}}$.

Next, we compute the derivative of the predicted acceleration $\mathbf{a}$ with respect to the parameter $\theta$. We start from Equation \eqref{A.2} and apply the product rule for differentiation:

\begin{align}
\frac{\partial \mathbf{a}}{\partial \theta} = \frac{\partial}{\partial \theta} \left( \mathbf{H}^{-1} \right) [\mathbf{g} - \mathbf{C}\dot{\mathbf{q}}] + \mathbf{H}^{-1} \frac{\partial}{\partial \theta} [\mathbf{g} - \mathbf{C}\dot{\mathbf{q}}] \label{A.5}
\end{align}

The crucial element in this expression is the derivative of the matrix inverse, $\mathbf{H}^{-1}$. The identity for the derivative of a matrix inverse $\mathbf{A}^{-1}$ with respect to a scalar $x$ is:
\begin{equation}
\frac{d \mathbf{A}^{-1}}{dx} = -\mathbf{A}^{-1} \frac{d\mathbf{A}}{dx} \mathbf{A}^{-1}
\end{equation}
Applying this identity to the Hessian matrix $\mathbf{H}$, we get:

\begin{align}
\frac{\partial \mathbf{H}^{-1}}{\partial \theta} = -\mathbf{H}^{-1} \frac{\partial \mathbf{H}}{\partial \theta} \mathbf{H}^{-1} \label{A.6}
\end{align}

The derivative of the second term in Equation \eqref{A.5} is more straightforward:
\begin{align}
\frac{\partial}{\partial \theta} [\mathbf{g} - \mathbf{C}\dot{\mathbf{q}}] = \frac{\partial \mathbf{g}}{\partial \theta} - \frac{\partial \mathbf{C}}{\partial \theta}\dot{\mathbf{q}} \label{A.7}
\end{align}

Substituting \eqref{A.6} and \eqref{A.7} back into \eqref{A.5} gives the full expression for the derivative of the acceleration:

\begin{align}
\frac{\partial \mathbf{a}}{\partial \theta} = \left( -\mathbf{H}^{-1} \frac{\partial \mathbf{H}}{\partial \theta} \mathbf{H}^{-1} \right) [\mathbf{g} - \mathbf{C}\dot{\mathbf{q}}] + \mathbf{H}^{-1} \left[ \frac{\partial \mathbf{g}}{\partial \theta} - \frac{\partial \mathbf{C}}{\partial \theta}\dot{\mathbf{q}} \right]
\end{align}

Recognizing that $[\mathbf{g} - \mathbf{C}\dot{\mathbf{q}}] = \mathbf{H}\mathbf{a}$ from Equation \eqref{A.3}, we can simplify the first term:

\begin{align}
\frac{\partial \mathbf{a}}{\partial \theta} &= \left( -\mathbf{H}^{-1} \frac{\partial \mathbf{H}}{\partial \theta} \mathbf{H}^{-1} \right) [\mathbf{H}\mathbf{a}] + \mathbf{H}^{-1} \left[ \frac{\partial \mathbf{g}}{\partial \theta} - \frac{\partial \mathbf{C}}{\partial \theta}\dot{\mathbf{q}} \right] \\
&= -\mathbf{H}^{-1} \frac{\partial \mathbf{H}}{\partial \theta} \mathbf{a} + \mathbf{H}^{-1} \left[ \frac{\partial \mathbf{g}}{\partial \theta} - \frac{\partial \mathbf{C}}{\partial \theta}\dot{\mathbf{q}} \right] \label{A.8}
\end{align}

Finally, we combine the results from \eqref{A.4} and \eqref{A.8} into Equation \eqref{A.3} to obtain the complete expression for the loss gradient with respect to the parameter $\theta$. The total gradient is the scalar product of the upstream gradient vector and the Jacobian of the acceleration:

\begin{align}
\frac{\partial \mathcal{L}_{\text{MAE}}}{\partial \theta} = \boldsymbol{\delta}_{\mathbf{a}}^{\top} \frac{\partial \mathbf{a}}{\partial \theta}
\end{align}

Substituting Equation \eqref{A.7}:

\begin{align}
\frac{\partial \mathcal{L}_{\text{MAE}}}{\partial \theta} = \boldsymbol{\delta}_{\mathbf{a}}^{\top} \left( -\mathbf{H}^{-1} \frac{\partial \mathbf{H}}{\partial \theta} \mathbf{a} + \mathbf{H}^{-1} \left[ \frac{\partial \mathbf{g}}{\partial \theta} - \frac{\partial \mathbf{C}}{\partial \theta}\dot{\mathbf{q}} \right] \right) \label{A.9}
\end{align}

This final expression confirms that the backpropagation pass requires the Hessian inverse $\mathbf{H}^{-1}$, which is also computed during the forward pass.

\begin{figure*}[htbp]
    \centering
    \includegraphics[width=\textwidth]{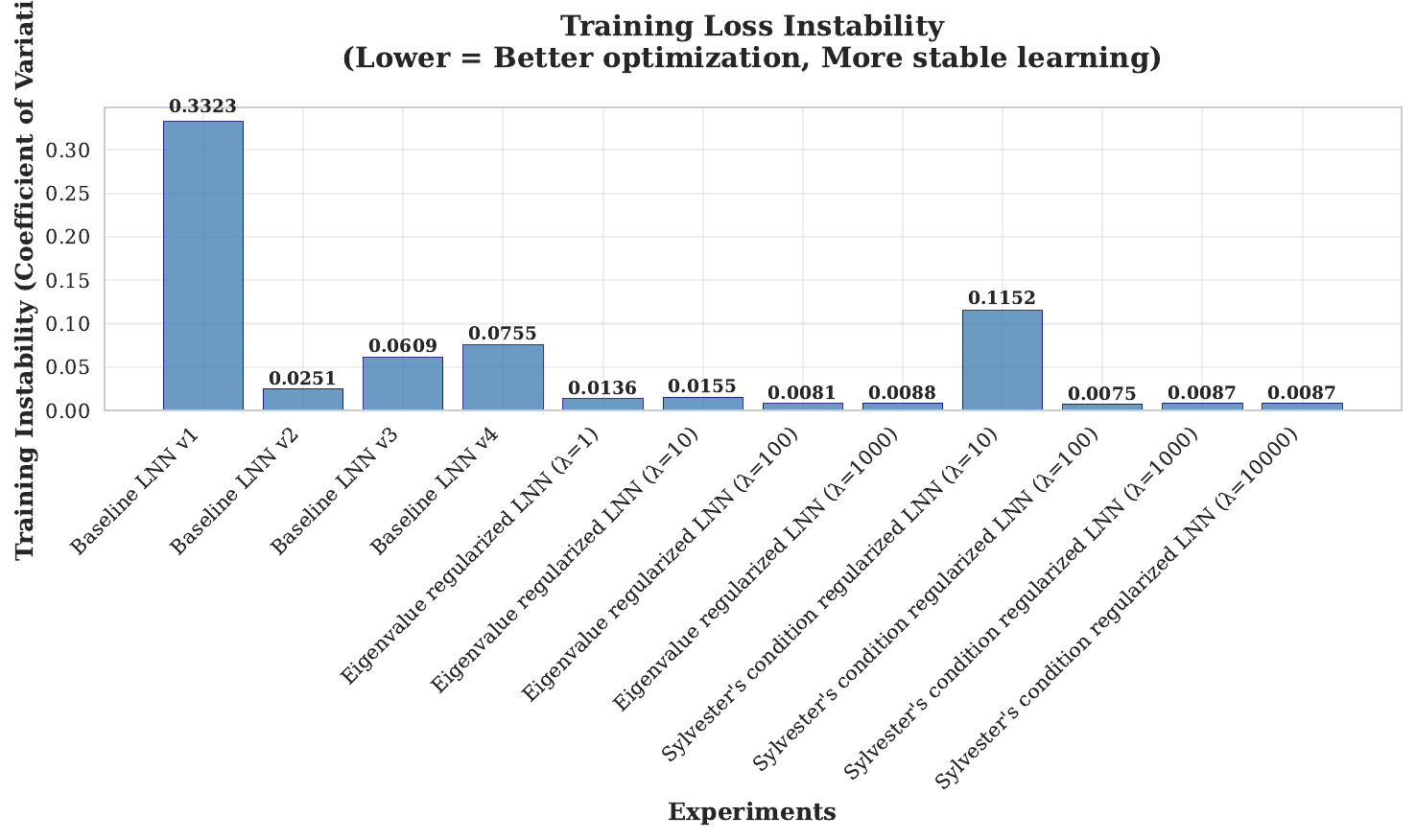}
    \caption{\capRegularizationComparison}
    \label{fig:regularization_comparison}
\end{figure*}

\section{Training Instability Definition}\label{Appendix:Instability}

We define training instability as the amplitude of high-frequency oscillations around a smoothed loss trend. To compute this, we first apply a Savitzky-Golay filter to the loss trajectory to extract the underlying trend. The filter uses polynomial order 2 and window size 15, which preserves the overall loss behavior while removing noise.

The oscillations are then calculated as the difference between the raw loss values and the smoothed trend. Since we use cosine annealing learning rate schedules, we analyze stability separately during the decreasing and increasing learning rate phases. For each phase, we compute the standard deviation of the oscillations.

The final stability metric is the average of these standard deviations across all learning rate cycles. Lower values indicate more stable training with fewer oscillations, while higher values suggest unstable optimization. When we report that stability improved by X\%, this means the average oscillation amplitude decreased by that percentage compared to the baseline.

This metric effectively captures training instabilities that manifest as erratic loss behavior, which is particularly important for LNNs where the unusual optimization landscape can lead to sudden spikes in the loss function.

\section{Training Details and Extra Figures of Network Experiments}\label{Appendix:TrainingDetailsNetwork}

All network experiments in this section use the double pendulum system described in Appendix~\ref{Appendix:DoublePendulumPhysical} (physical constants, Lagrangian, and analytical acceleration equations). Unless otherwise specified, experiments use the following \textbf{baseline configuration}:

\begin{itemize}
\item \textbf{Training:} 60,000 samples (validation: 12,000); position bounds $\theta_i \in [-3\pi, 3\pi]$; velocity bounds $\dot{\theta}_i \in [-10, 10]$ rad/s; physical scaling $s_{\theta_i} = 10.00$
\item \textbf{Optimization:} Adam (weight decay $10^{-6}$); lr $= 10^{-3}$; 300 epochs; batch size 128; Cosine Annealing ($T_{\max} = 30$, $\eta_{\min} = 10^{-6}$)
\item \textbf{Architecture:} 4-layer MLP; 500 hidden units; GeLU activation; $+1 \cdot \lVert\dot{q}\rVert^2$ term
\item \textbf{Trajectory testing:} 4 random initial conditions with $\theta_i \in [0, 2\pi]$, $\dot{\theta}_i \in [-0.1, 0.1]$ rad/s, $t \in [0, 100]$ s ($\Delta t = 0.1$ s); ground truth via LSODA~\cite{petzold1983} (rtol/atol $= 10^{-10}$); LNN via PIDController~\cite{lienen2022} (rtol $= 10^{-6}$, atol $= 10^{-8}$, $p_{\text{coeff}} = 0.2$, $i_{\text{coeff}} = 0.4$, $d_{\text{coeff}} = 0$)
\end{itemize}

Each subsection below states any deviations from this baseline; unlisted parameters match the baseline.

Also, it should be noted that since sometimes the training dataset had wider bounds detailed in \ref{subsec:training_bounds}, the training error was bigger than the validation loss, which might seem unusual at first.

\subsection{Effect of the Hessian Regularization}\label{Appendix:HessianRegularization}

\begin{figure*}[htbp]
    \centering
    \includegraphics[width=\textwidth]{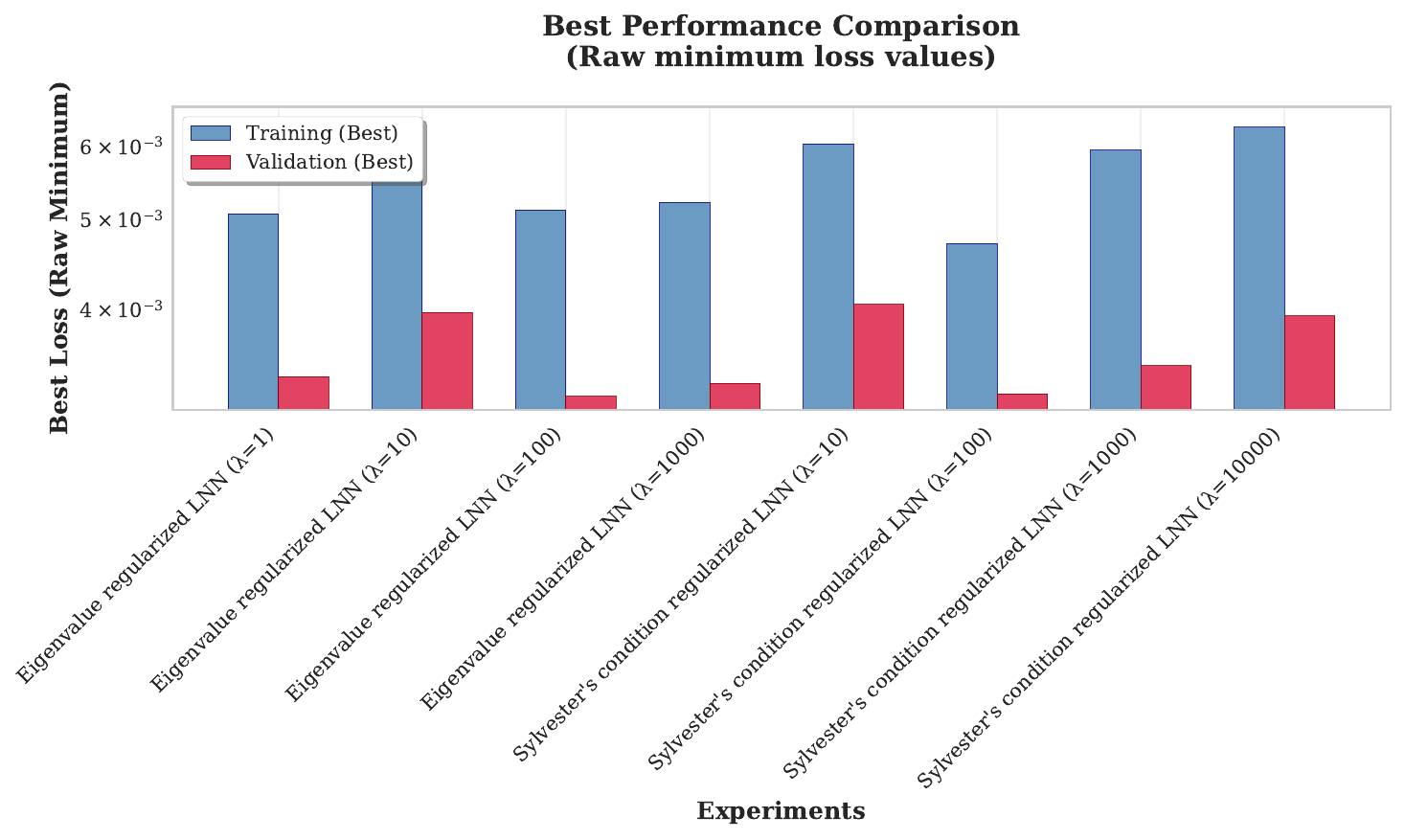}
    \caption{\capLambdaValues}
    \label{fig:lambda_values}
\end{figure*}

\noindent\textbf{Deviations from baseline:} Position bounds $\theta_i \in [-\pi/2, 5\pi/2]$ rad.

\noindent\textbf{Matches baseline:} Physical system, velocity bounds, Training data, epochs, optimizer, scheduler, trajectory test settings, network architecture (4-layer MLP, 500 hidden, GeLU, $+\|\dot{q}\|^2$).

\vspace{0.3cm}
\noindent\textbf{Regularization comparison}

\begin{tabular}{lll}
\hline
\textbf{Method} & \textbf{$\lambda$} & \textbf{Runs} \\
\hline
Baseline LNN & 0 & 4$^\dagger$ \\
Eigenvalue reg. & 1, 10, 100, 1000 & 1 each \\
Sylvester reg. & 10, 100, 1000, 10000 & 1 each \\
\hline
\end{tabular}

\vspace{0.1cm}
{\footnotesize $^\dagger$Different random seeds for network weight initialization.}

\subsection{Effect of the Normalization Method}\label{Appendix:NormalizationComparison}

\noindent\textbf{Deviations from baseline:} Activation $x\tanh(kx)$ instead of GeLU; physical scaling varies by method (see table).

\noindent\textbf{Matches baseline:} Physical system, position bounds $[-3\pi, 3\pi]$, velocity bounds, Training data, epochs, optimizer, scheduler, trajectory test settings, architecture (4-layer MLP, 500 hidden, $+\|\dot{q}\|^2$).

\vspace{0.3cm}
\noindent\textbf{Normalization comparison}

\begin{tabular}{ll}
\hline
\textbf{Method} & \textbf{Scale} \\
\hline
LNN with DyT layer & $s_{\theta_i} = 1.00$ \\
LNN with Physical Scaling & $s_{\theta_i} = 10.00$ \\
LNN without Normalization & $s_{\theta_i} = 1.00$ \\
\hline
\end{tabular}

\subsection{Effect of the Initial Quadratic Term}\label{Appendix:ExtraTermComparison}

\noindent\textbf{Deviations from baseline:} No $+\|\dot{q}\|^2$ term by default (compared with and without); activation varies per run (see table).

\noindent\textbf{Matches baseline:} Physical system, physical scaling ($s_{\theta_i} = 10$), position bounds $[-3\pi, 3\pi]$, velocity bounds, Training data, epochs, optimizer, scheduler, trajectory test settings, architecture (4-layer MLP, 500 hidden).

\vspace{0.3cm}
\noindent\textbf{Extra term comparison}

\begin{tabular}{lll}
\hline
\textbf{Activation} & \textbf{Extra Term} \\
\hline
GeLU & Yes ($+ 1 \cdot \|\dot{q}\|^2$) \\
GeLU & No \\
$x\tanh(kx)$ & Yes ($+ 1 \cdot \|\dot{q}\|^2$) \\
$x\tanh(kx)$ & No \\
Softplus & Yes ($+ 1 \cdot \|\dot{q}\|^2$) \\
Softplus & No \\
\hline
\end{tabular}

\subsection{Effect of Activation Functions}\label{Appendix:ActivationComparison}

\begin{figure*}[htbp]
\centering
\includegraphics[width=\textwidth]{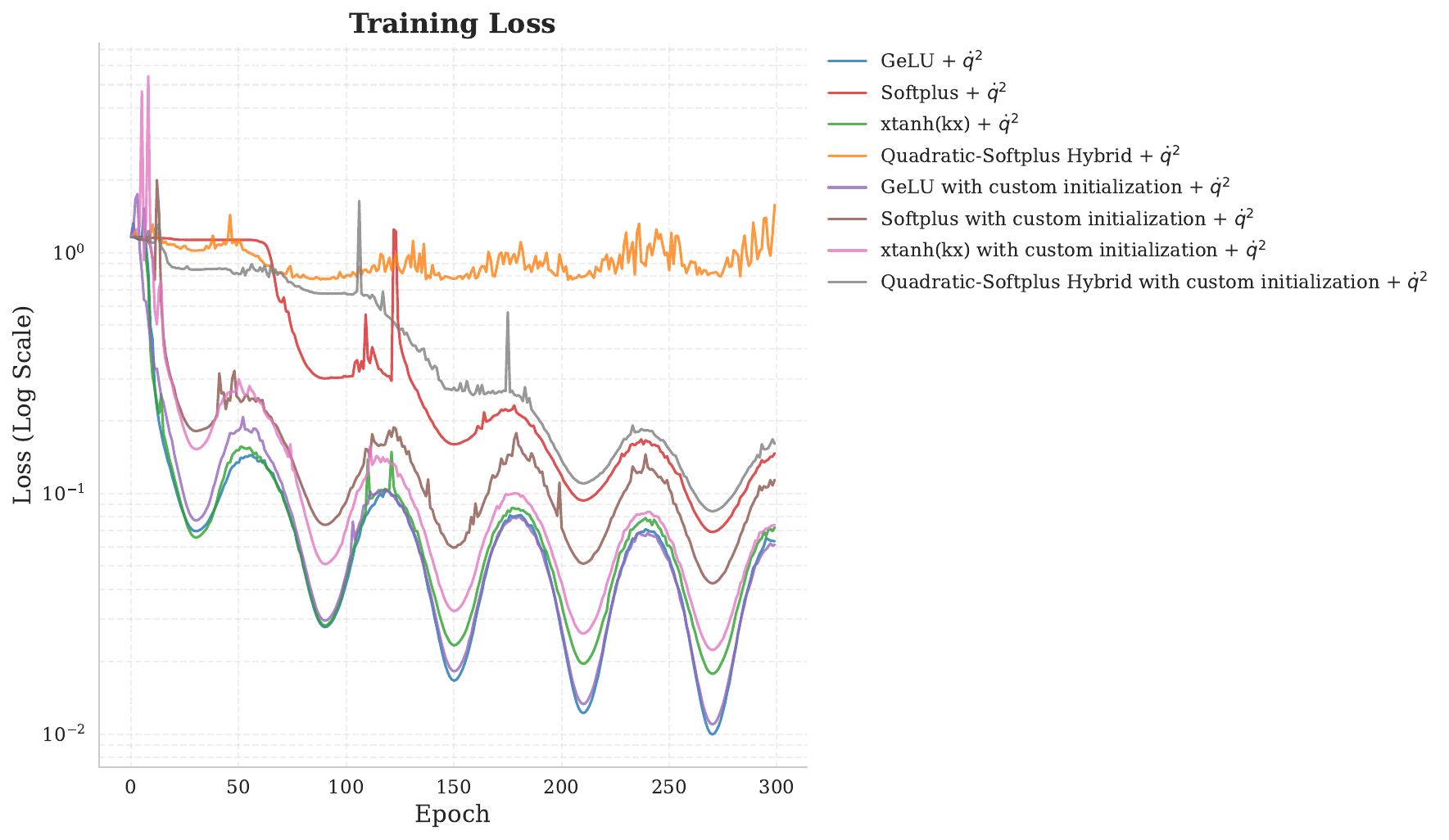}
\caption{\capActivationComparison}
\label{fig:activation_comparison}
\end{figure*}

\noindent\textbf{Deviations from baseline:} Activation and initialization vary (see table); Hybrid architecture uses hidden dimension $250 + 250$.

\noindent\textbf{Matches baseline:} Physical system, physical scaling ($s_{\theta_i} = 10$), position bounds $[-3\pi, 3\pi]$, velocity bounds, Training data, epochs, optimizer, scheduler, trajectory test settings, 4-layer MLP with $+\|\dot{q}\|^2$.

\vspace{0.1cm}
{\footnotesize $^\dagger$Hybrid architecture uses two parallel networks: a Softplus MLP and a Quadratic MLP (QMLP) where the output passes through $x \mapsto x^2$. Each network has hidden dimension 250.}

\vspace{3cm}
\noindent\textbf{Activation and initialization comparison}

\begin{tabular}{ll}
\hline
\textbf{Activation} & \textbf{Initialization} \\
\hline
GeLU & Default \\
GeLU & Custom \\
Softplus & Default \\
Softplus & Custom \\
$x\tanh(kx)$ & Default \\
$x\tanh(kx)$ & Custom \\
Quadratic-Softplus Hybrid$^\dagger$ & Default \\
Quadratic-Softplus Hybrid$^\dagger$ & Custom \\
\hline
\end{tabular}

\subsection{Effect of Weight Initialization}\label{Appendix:InitializationComparison}

\noindent\textbf{Deviations from baseline:} Initialization and activation vary (see table); Hybrid architecture uses hidden dimension $250 + 250$.

\noindent\textbf{Matches baseline:} Physical system, physical scaling ($s_{\theta_i} = 10$), position bounds $[-3\pi, 3\pi]$, velocity bounds, Training data, epochs, optimizer, scheduler, trajectory test settings, 4-layer MLP with $+\|\dot{q}\|^2$.

\vspace{0.1cm}
{\footnotesize $^\dagger$Hybrid architecture uses two parallel networks: a Softplus MLP and a Quadratic MLP (QMLP) where the output passes through $x \mapsto x^2$. Each network has hidden dimension 250.}

\vspace{0.3cm}
\noindent\textbf{Initialization comparison}

\begin{tabular}{lll}
\hline
\textbf{Initialization} & \textbf{Activation} & \textbf{Runs} \\
\hline
Default & GeLU & 2$^\ddagger$ \\
Default & Softplus & 1 \\
Default & $x\tanh(kx)$ & 1 \\
Default & Quadratic-Softplus Hybrid$^\dagger$ & 1 \\
Custom & GeLU & 2$^\ddagger$ \\
Custom & Softplus & 1 \\
Custom & $x\tanh(kx)$ & 1 \\
Custom & Quadratic-Softplus Hybrid$^\dagger$ & 1 \\
\hline
\end{tabular}

\begin{figure*}[htbp]
\centering
\includegraphics[width=\textwidth]{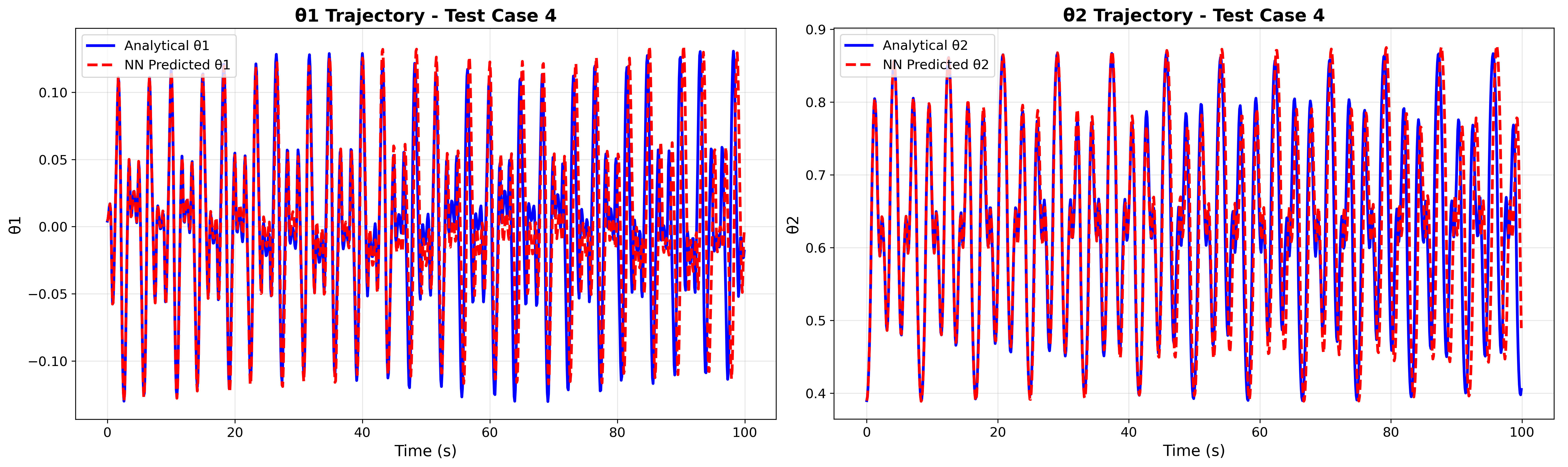}
\caption{\capDoublePendulumCoordError}
\label{fig:double_pendulum_coord_error}
\end{figure*}

\vspace{0.1cm}
{\footnotesize $^\ddagger$Different random seeds for network weight initialization (v1, v2).}

\subsection{Effect of Training Range}\label{Appendix:TrainingRangeComparison}

\noindent\textbf{Deviations from baseline:} Position bounds vary (see table).

\noindent\textbf{Matches baseline:} Physical system, physical scaling ($s_{\theta_i} = 10$), velocity bounds, Training data, epochs, optimizer, scheduler, trajectory test settings, architecture (4-layer MLP, 500 hidden, GeLU, $+\|\dot{q}\|^2$).

\vspace{0.3cm}
\noindent\textbf{Training range comparison}

\begin{tabular}{ll}
\hline
\textbf{Position Bounds} & \textbf{Runs} \\
\hline
$\theta_i \in [-3\pi, 3\pi]$ & 1 \\
$\theta_i \in [0, 2\pi]$ & 2$^\dagger$ \\
$\theta_i \in [-\pi/2, 2\pi + \pi/2]$ & 1 \\
$\theta_i \in [-\pi/4, 2\pi + \pi/4]$ & 1 \\
$\theta_i \in [-\pi/10, 2\pi + \pi/10]$ & 1 \\
\hline
\end{tabular}

\vspace{0.1cm}
{\footnotesize $^\dagger$Different random seeds for network weight initialization (v1, v2).}
\subsection{Comparison with the Previously Suggested Tricks}\label{Appendix:OldPapersComparison}

\newcommand{\CranmerInit}{Cranmer Init.}
\newcommand{\LiuTricks}{Liu (Tricks)}
\newcommand{\Ours}{Ours}

\noindent\textbf{Shorthand notation:} For compactness in the tables below we denote the \citet{cranmerLagrangianNeuralNetworks2020} initialized model that we trained by \textbf{\CranmerInit{}} and we denote the model with the tricks proposed by \citet{Liu:2021xwt} applied by \textbf{\LiuTricks{}}. Full citations are kept elsewhere in the manuscript.

\noindent\textbf{Deviations from baseline:} Architecture, activation, initialization, regularization, and position bounds vary by model (see tables below).

\noindent\textbf{Matches baseline:} Physical system, physical scaling ($s_{\theta_i} = 10$), velocity bounds, Training data, epochs, optimizer, scheduler, trajectory test settings.

\vspace{0.3cm}
\noindent\textbf{Training position bounds comparison}

\begin{tabular}{ll}
\hline
\textbf{Model} & \textbf{Position Bounds} \\
\hline
\CranmerInit{} & $\theta_i \in [0, 2\pi]$ \\
\LiuTricks{} & $\theta_i \in [0, 2\pi]$ \\
\Ours{} ($\lambda = 100$) & $\theta_i \in [-\pi/2, 2\pi + \pi/2]$ \\
\hline
\end{tabular}

\vspace{0.3cm}
\noindent\textbf{Architecture comparison}

{\small
\begin{tabular}{lccc}
\hline
& \makecell{\textbf{\CranmerInit{}}} & \makecell{\textbf{\LiuTricks{}}} & \textbf{\Ours{}} \\
\hline
Type & MLP & MLP + QMLP$^\dagger$ & MLP \\
Layers & 4 & 4 & 4 \\
Hidden dim & 500 & 250 + 250 & 500 \\
Activation & Softplus & Softplus \& $x^2$ & GeLU \\
$\|\dot{q}\|^2$ term & No & Yes & Yes \\
Initialization & Cranmer$^\ddagger$ & Default & Default \\
Reg.\ $\lambda$ & 0 & 0 & 100 \\
\hline
\end{tabular}
}

\vspace{0.1cm}
{\footnotesize $^\dagger$Hybrid architecture uses two parallel networks: a Softplus MLP and a Quadratic MLP (QMLP) where the output passes through $x \mapsto x^2$. Each network has hidden dimension 250.}

\vspace{0.1cm}
{\footnotesize $^\ddagger$Cranmer initialization~\cite{cranmerLagrangianNeuralNetworks2020}: weights initialized with $\mathcal{N}(0, \sigma^2)$ where $\sigma = v_i / \sqrt{n}$, with $v_1 = 2.2$ for the first layer, $v_i = 0.58 \cdot i$ for hidden layer $i$, and $v_L = n$ for the output layer.}

\subsection{Double Pendulum}

\section{Training Details and Extra Figures of Physical Experiments}\label{Appendix:TrainingDetailsPhysical}

\subsection{Double Pendulum}\label{Appendix:DoublePendulumPhysical}

\subsubsection*{Physical Settings}

\begin{tabular}{ll}
\textbf{Inputs} & $[\theta_1, \theta_2, \dot{\theta}_1, \dot{\theta}_2]$ \\
\textbf{Physical scaling} & $s_{\theta_1} = 10.00$, $s_{\theta_2} = 10.00$ \\
\textbf{Constants} & $m_1 = 1$ kg, $m_2 = 1$ kg, $l_1 = 1$ m, \\
& $l_2 = 1$ m, $g = 9.8$ m/s$^2$ \\
\end{tabular}

\vspace{0.3cm}
\noindent\textbf{Lagrangian}
\begin{align*}
L(\theta, \dot{\theta}) &= \frac{1}{2}m_1(l_1\dot{\theta}_1)^2 + \frac{1}{2}m_2\bigl[(l_1\dot{\theta}_1)^2 + (l_2\dot{\theta}_2)^2 \\
&\quad + 2l_1 l_2 \dot{\theta}_1\dot{\theta}_2 \cos(\theta_1 - \theta_2)\bigr] \\
&\quad + m_1 g l_1 \cos(\theta_1) + m_2 g[l_1\cos(\theta_1) + l_2\cos(\theta_2)]
\end{align*}

\noindent\textbf{Analytical acceleration}
\begin{align*}
a_1 &= \frac{l_2}{l_1} \cdot \frac{m_2}{m_1 + m_2} \cos(\theta_1 - \theta_2), \quad
a_2 = \frac{l_1}{l_2} \cos(\theta_1 - \theta_2) \\
f_1 &= -\frac{l_2}{l_1} \cdot \frac{m_2}{m_1 + m_2} \dot{\theta}_2^2 \sin(\theta_1 - \theta_2) - \frac{g}{l_1}\sin(\theta_1) \\
f_2 &= \frac{l_1}{l_2} \dot{\theta}_1^2 \sin(\theta_1 - \theta_2) - \frac{g}{l_2}\sin(\theta_2) \\
\ddot{\theta}_1 &= \frac{f_1 - a_1 f_2}{1 - a_1 a_2}, \quad
\ddot{\theta}_2 = \frac{f_2 - a_2 f_1}{1 - a_1 a_2}
\end{align*}

\begin{figure*}[htbp]
\centering
\includegraphics[width=\textwidth]{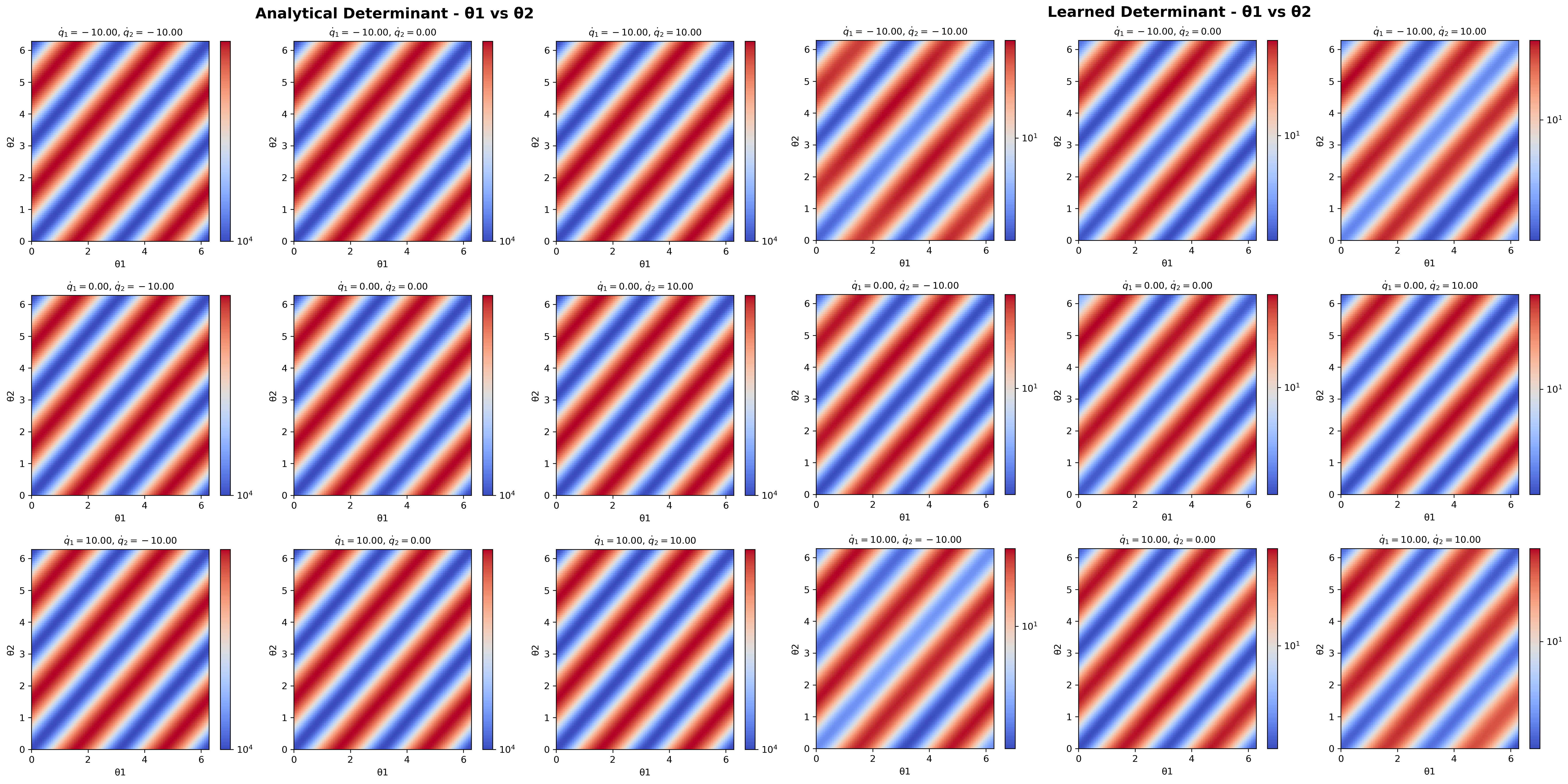}
\caption{\capDoublePendulumHessian}
\label{fig:double_pendulum_hessian}
\end{figure*}

\subsubsection*{Training Settings}

Training was performed in three sequential phases, with each phase initialized from the best model of the previous phase. Total training: 700 epochs.

\begin{table}
\squeezetable
\begin{ruledtabular}
\begin{tabular}{ll}
\textbf{Training data} & 60,000 uniformly random samples \\
\textbf{Velocity bounds} & $\dot{\theta}_i \in [-10, 10]$ rad/s \\
\textbf{Validation data} & 12,000 (20\% of training) \\
\textbf{Mini-batch size} & 128 \\
\textbf{Optimizer} & Adam with weight decay = $10^{-6}$ \\
\textbf{Scheduler} & Cosine Annealing, $T_{\max} = 30$ \\
\end{tabular}
\end{ruledtabular}
\caption{Training settings.}
\end{table}

\begin{figure}[htbp]
\centering
\includegraphics[width=0.48\textwidth]{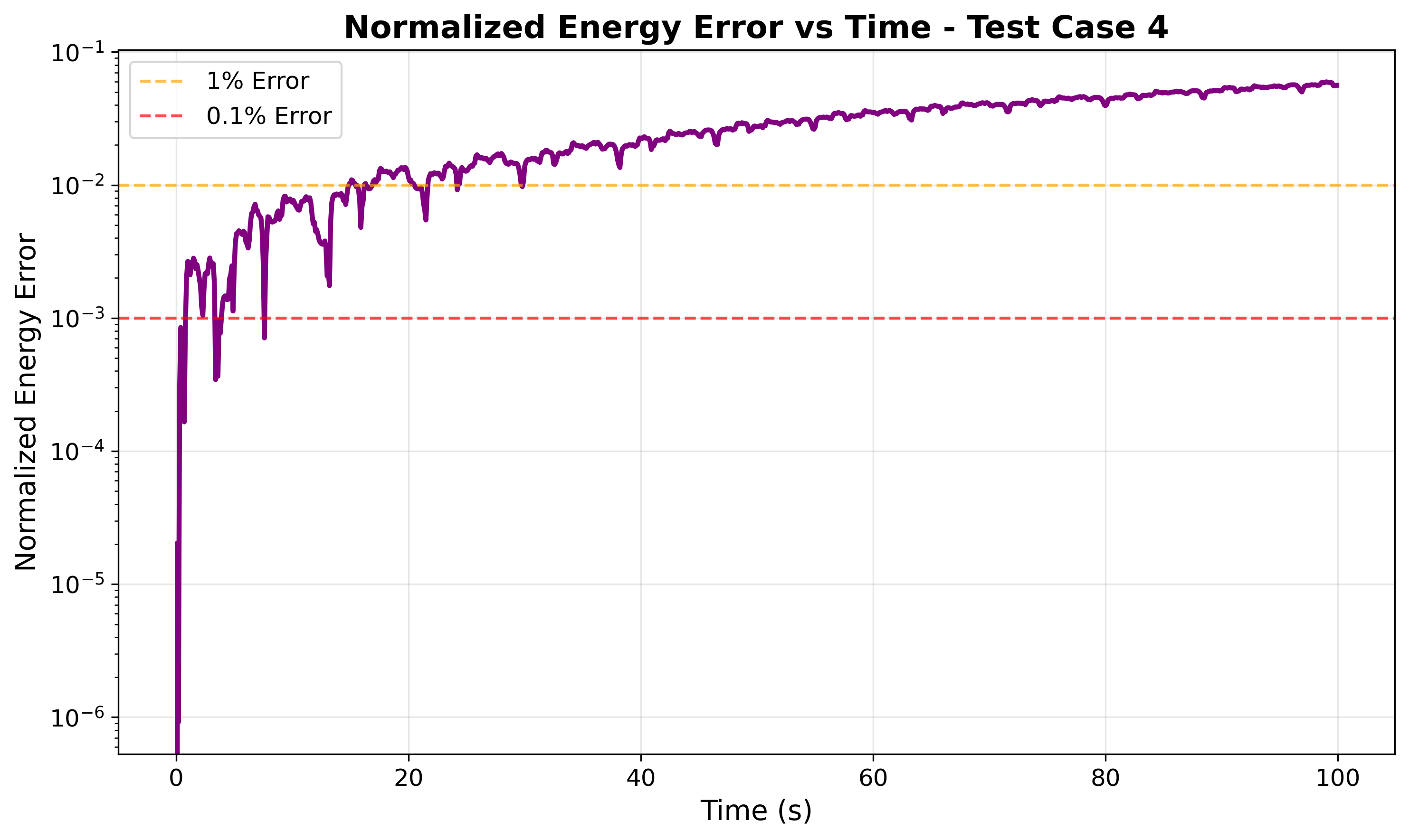}
\caption{\capDoublePendulumEnergy}
\label{fig:double_pendulum_energy}
\end{figure}

\vspace{0.3cm}
\noindent\textbf{Multi-phase training schedule}
{\small
\begin{tabular}{lccccc}
\hline
Phase & Epochs & Total Eps. & lr & $\eta_{\min}$ & Position Bounds \\
\hline
1 & 300 & 300 & $10^{-3}$ & $10^{-6}$ & $[-3\pi, 3\pi]$ \\
2 & 200 & 500 & $4{\times}10^{-5}$ & $10^{-7}$ & $[-\frac{\pi}{10}, \frac{21\pi}{10}]$ \\
3 & 200 & 700 & $10^{-5}$ & $10^{-8}$ & $[-\frac{\pi}{10}, \frac{21\pi}{10}]$ \\
\hline
\end{tabular}
}

\begin{figure}[htbp]
\centering
\includegraphics[width=0.48\textwidth]{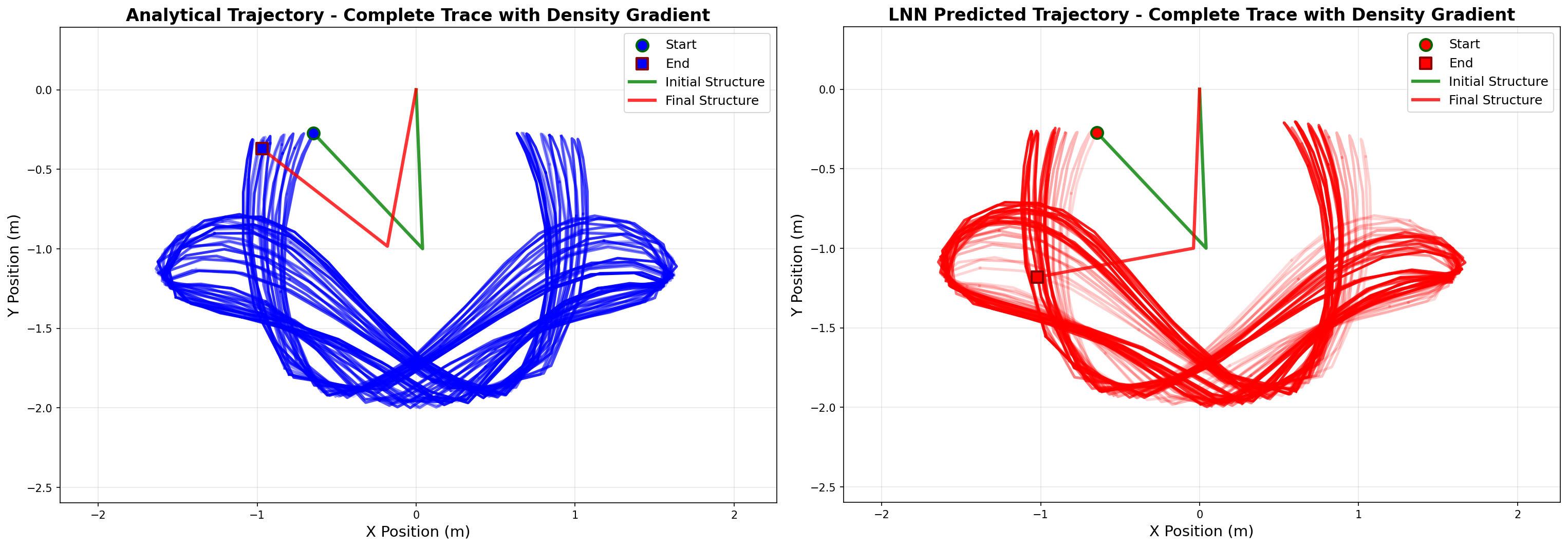}
\caption{\capDoublePendulumTrace}
\label{fig:double_pendulum_trace}
\end{figure}

\subsubsection*{Trajectory Test Settings}

\squeezetable
\begin{ruledtabular}
\begin{tabular}{ll}
\textbf{Initial conditions} & 4 uniformly random samples \\
\textbf{Position bounds} & $\theta_i \in [0, 2\pi]$ rad \\
\textbf{Velocity bounds} & $\dot{\theta}_i \in [-0.1, 0.1]$ rad/s \\
\textbf{Time span} & $t \in [0, 100]$ s with $\Delta t = 0.1$ s \\
\textbf{Baseline solver} & LSODA~\cite{petzold1983} \\ & (rtol=$10^{-10}$, atol=$10^{-10}$) \\
\textbf{LNN solver} & PIDController~\cite{lienen2022} (rtol=$10^{-6}$, \\
& atol=$10^{-8}$, $p_{\text{coeff}}=0.2$, \\
& $i_{\text{coeff}}=0.4$, $d_{\text{coeff}}=0$) \\
\end{tabular}
\end{ruledtabular}

\subsubsection*{Network Architecture}

\squeezetable
\begin{ruledtabular}
\begin{tabular}{ll}
\textbf{Type} & MLP $+ 1 \cdot \|\dot{q}\|^2$ \\
\textbf{Layers} & 4 \\
\textbf{Hidden dimension} & 500 \\
\textbf{Activation} & GeLU \\
\textbf{Regularization} & Eigenvalue, $\lambda = 5$ \\
\end{tabular}
\end{ruledtabular}

\subsection{Spring Pendulum}\label{Appendix:SpringPendulumPhysical}

\subsubsection*{Physical Settings}

\squeezetable
\begin{ruledtabular}
\begin{tabular}{ll}
\textbf{Inputs} & $[r, \theta, \dot{r}, \dot{\theta}]$ \\
\textbf{Physical scaling} & $s_r = 5.00$, $s_\theta = 7.85$ \\
\textbf{Constants} & $m = 1$ kg, $k = 40$ N/m, \\
& $g = 9.8$ m/s$^2$, $\ell_0 = 1$ m \\
\end{tabular}
\end{ruledtabular}

\vspace{0.3cm}
\noindent\textbf{Lagrangian}
\begin{align*}
L(r, \theta, \dot{r}, \dot{\theta}) &= \frac{1}{2}m(\dot{r}^2 + r^2\dot{\theta}^2) + mgr\cos\theta - \frac{1}{2}k(r - \ell_0)^2
\end{align*}

\noindent\textbf{Analytical acceleration}
\begin{align*}
\ddot{r} &= r\dot{\theta}^2 + g\cos\theta - \frac{k}{m}(r - \ell_0) \\
\ddot{\theta} &= \frac{-g\sin\theta - 2\dot{r}\dot{\theta}}{r}
\end{align*}

\subsubsection*{Training Settings}

\squeezetable
\begin{ruledtabular}
\begin{tabular}{ll}
\textbf{Training data} & 60,000 uniformly random samples \\
\textbf{Position bounds} & $r \in [0.05, 5]$ m, \\
& $\theta \in [-\pi/2, 5\pi/2]$ rad \\
\textbf{Velocity bounds} & $\dot{r} \in [-5, 5]$ m/s, \\
& $\dot{\theta} \in [-5, 5]$ rad/s \\
\textbf{Validation data} & 12,000 (20\% of training) \\
\textbf{Epochs} & 500 \\
\textbf{Mini-batch size} & 128 \\
\textbf{Learning rate} & $5 \times 10^{-4}$ \\
\textbf{Optimizer} & Adam with weight decay = $10^{-6}$ \\
\textbf{Scheduler} & Cosine Annealing, $T_{\max} = 30$, \\
& $\eta_{\min} = 10^{-7}$ \\
\end{tabular}
\end{ruledtabular}

\subsubsection*{Trajectory Test Settings}

\squeezetable
\begin{ruledtabular}
\begin{tabular}{ll}
\textbf{Initial conditions} & 4 uniformly random samples \\
\textbf{Position bounds} & $r \in [0.5, 1.5]$ m, $\theta \in [0, 2\pi]$ rad \\
\textbf{Velocity bounds} & $\dot{r} \in [-1, 1]$ m/s, $\dot{\theta} \in [-1, 1]$ rad/s \\
\textbf{Time span} & $t \in [0, 50]$ s with $\Delta t = 0.1$ s \\
\textbf{Baseline solver} & LSODA~\cite{petzold1983} \\
& (rtol=$10^{-10}$, atol=$10^{-10}$) \\
\textbf{LNN solver} & PIDController~\cite{lienen2022} (rtol=$10^{-6}$, \\
& atol=$10^{-8}$, $p_{\text{coeff}}=0.2$, \\
& $i_{\text{coeff}}=0.4$, $d_{\text{coeff}}=0$) \\
\end{tabular}
\end{ruledtabular}

\subsubsection*{Network Architecture}

\squeezetable
\begin{ruledtabular}
\begin{tabular}{ll}
\textbf{Type} & MLP $+ 1 \cdot \|\dot{q}\|^2$ \\
\textbf{Layers} & 3 \\
\textbf{Hidden dimension} & 600 \\
\textbf{Activation} & GeLU \\
\textbf{Regularization} & Eigenvalue, $\lambda = 5$ \\
\end{tabular}
\end{ruledtabular}

\FloatBarrier

\subsection{Triple Pendulum}\label{Appendix:TriplePendulumPhysical}

\begin{figure}[htbp]
\centering
\includegraphics[width=0.48\textwidth]{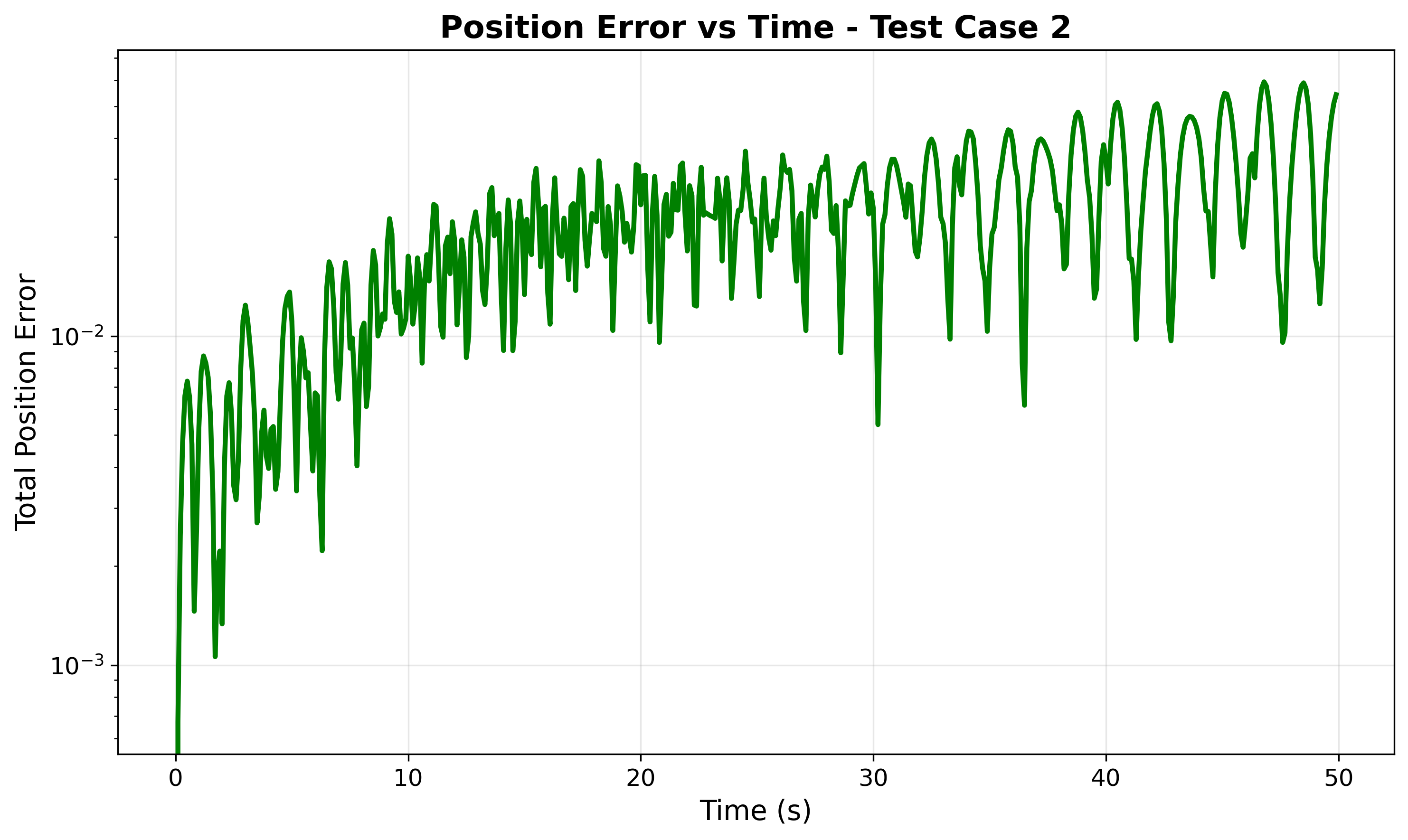}
\caption{\capTriplePendulumCoordErrorNonChaotic}
\label{fig:triple_pendulum_coord_error_nonchaotic}
\end{figure}

\subsubsection*{Physical Settings}

\squeezetable
\begin{ruledtabular}
\begin{tabular}{ll}
\textbf{Inputs} & $[\theta_1, \theta_2, \theta_3, \dot{\theta}_1, \dot{\theta}_2, \dot{\theta}_3]$ \\
\textbf{Physical scaling} & $s_{\theta_1} = s_{\theta_2} = s_{\theta_3} = 10.00$ \\
\textbf{Constants} & $m_1 = m_2 = m_3 = 1$ kg, \\
& $l_1 = l_2 = l_3 = 1$ m, $g = 9.8$ m/s$^2$ \\
\end{tabular}
\end{ruledtabular}

\begin{figure*}[htbp]
\centering
\includegraphics[width=\textwidth]{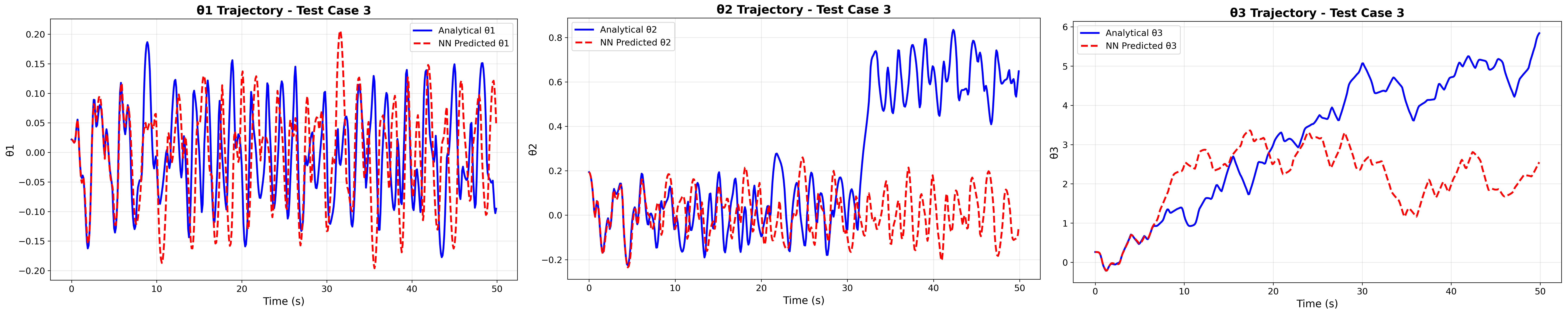}
\caption{\capTriplePendulumCoordErrorChaotic}
\label{fig:triple_pendulum_coord_error_chaotic}
\end{figure*}

\vspace{0.3cm}
\noindent\textbf{Lagrangian}
{\small
\begin{align*}
T &= \frac{1}{2}(m_1 + m_2 + m_3) l_1^2 \dot{\theta}_1^2 + \frac{1}{2}(m_2 + m_3) l_2^2 \dot{\theta}_2^2 + \frac{1}{2} m_3 l_3^2 \dot{\theta}_3^2 \\
&\quad + (m_2 + m_3) l_1 l_2 \dot{\theta}_1 \dot{\theta}_2 \cos(\theta_1 - \theta_2) \\
&\quad + m_3 l_1 l_3 \dot{\theta}_1 \dot{\theta}_3 \cos(\theta_1 - \theta_3) \\
&\quad + m_3 l_2 l_3 \dot{\theta}_2 \dot{\theta}_3 \cos(\theta_2 - \theta_3) \\
V &= -g \bigl[ (m_1 + m_2 + m_3) l_1 \cos\theta_1 \\
&\quad + (m_2 + m_3) l_2 \cos\theta_2 + m_3 l_3 \cos\theta_3 \bigr] \\
L &= T - V
\end{align*}
}

\begin{figure}[htbp]
\centering
\includegraphics[width=0.48\textwidth]{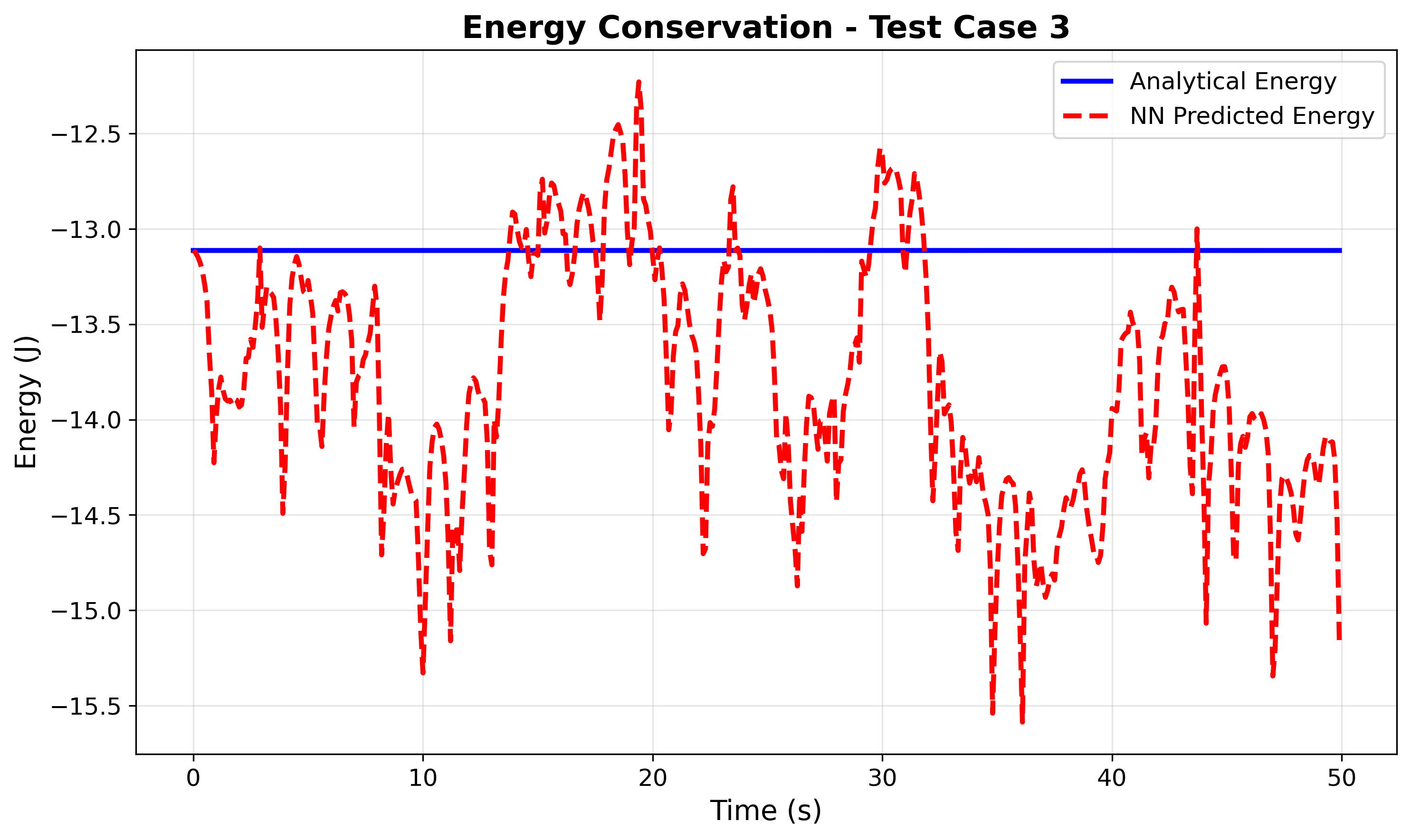}
\caption{\capTriplePendulumEnergyChaotic}
\label{fig:triple_pendulum_energy_chaotic}
\end{figure}

\noindent\textbf{Analytical acceleration}

The equations of motion are solved via $\mathbf{M}(\boldsymbol{\theta}) \ddot{\boldsymbol{\theta}} = \mathbf{F}(\boldsymbol{\theta}, \dot{\boldsymbol{\theta}})$, where the symmetric mass matrix is:
{\small
\begin{align*}
M_{11} &= (m_1 + m_2 + m_3) l_1^2, \quad M_{22} = (m_2 + m_3) l_2^2 \\
M_{33} &= m_3 l_3^2, \quad M_{12} = (m_2 + m_3) l_1 l_2 \cos(\theta_1 - \theta_2) \\
M_{13} &= m_3 l_1 l_3 \cos(\theta_1 - \theta_3) \\
M_{23} &= m_3 l_2 l_3 \cos(\theta_2 - \theta_3)
\end{align*}
}
and the force vector components are:
{\small
\begin{align*}
F_1 &= -(m_2 + m_3) l_1 l_2 \dot{\theta}_2^2 \sin(\theta_1 - \theta_2) \\
&\quad - m_3 l_1 l_3 \dot{\theta}_3^2 \sin(\theta_1 - \theta_3) - (m_1 + m_2 + m_3) g l_1 \sin\theta_1 \\
F_2 &= (m_2 + m_3) l_1 l_2 \dot{\theta}_1^2 \sin(\theta_1 - \theta_2) \\
&\quad - m_3 l_2 l_3 \dot{\theta}_3^2 \sin(\theta_2 - \theta_3) - (m_2 + m_3) g l_2 \sin\theta_2 \\
F_3 &= m_3 l_1 l_3 \dot{\theta}_1^2 \sin(\theta_1 - \theta_3) \\
&\quad + m_3 l_2 l_3 \dot{\theta}_2^2 \sin(\theta_2 - \theta_3) - m_3 g l_3 \sin\theta_3
\end{align*}
}

\subsubsection*{Training Settings}

\squeezetable
\begin{ruledtabular}
\begin{tabular}{ll}
\textbf{Training data} & 60,000 uniformly random samples \\
\textbf{Position bounds} & $\theta_i \in [-\pi/2, 5\pi/2]$ rad \\
\textbf{Velocity bounds} & $\dot{\theta}_i \in [-10, 10]$ rad/s \\
\textbf{Validation data} & 12,000 (20\% of training) \\
\textbf{Epochs} & 500 \\
\textbf{Mini-batch size} & 128 \\
\textbf{Learning rate} & $10^{-3}$ \\
\textbf{Optimizer} & Adam with weight decay = $10^{-6}$ \\
\textbf{Scheduler} & Cosine Annealing, $T_{\max} = 30$, \\
& $\eta_{\min} = 10^{-7}$ \\
\end{tabular}
\end{ruledtabular}

\subsubsection*{Trajectory Test Settings}

\squeezetable
\begin{ruledtabular}
\begin{tabular}{ll}
\textbf{Initial conditions} & 4 uniformly random samples \\
\textbf{Position bounds} & $\theta_i \in [0.2, 3.0]$ rad \\
\textbf{Velocity bounds} & $\dot{\theta}_i \in [-0.1, 0.1]$ rad/s \\
\textbf{Time span} & $t \in [0, 50]$ s with $\Delta t = 0.1$ s \\
\textbf{Baseline solver} & LSODA~\cite{petzold1983} \\
& (rtol=$10^{-10}$, atol=$10^{-10}$) \\
\textbf{LNN solver} & PIDController~\cite{lienen2022} (rtol=$10^{-6}$, \\
& atol=$10^{-8}$, $p_{\text{coeff}}=0.2$, \\
& $i_{\text{coeff}}=0.4$, $d_{\text{coeff}}=0$) \\
\end{tabular}
\end{ruledtabular}

\subsubsection*{Network Architecture}

\squeezetable
\begin{ruledtabular}
\begin{tabular}{ll}
\textbf{Type} & MLP $+ 1 \cdot \|\dot{q}\|^2$ \\
\textbf{Layers} & 4 \\
\textbf{Hidden dimension} & 500 \\
\textbf{Activation} & GeLU \\
\textbf{Regularization} & Eigenvalue, $\lambda = 5$ \\
\end{tabular}
\end{ruledtabular}

\FloatBarrier

\subsection{Sphere Geodesic}\label{Appendix:SphereGeodesicPhysical}

\subsubsection*{Physical Settings}

\squeezetable
\begin{ruledtabular}
\begin{tabular}{ll}
\textbf{Inputs} & $[\theta, \phi, \dot{\theta}, \dot{\phi}]$ \\
\textbf{Physical scaling} & $s_\theta = 5.00$, $s_\phi = 7.85$ \\
\textbf{Constants} & $m = 1$ kg, $R = 1$ m \\
\end{tabular}
\end{ruledtabular}

\begin{figure}[htbp]
\centering
\includegraphics[width=0.48\textwidth]{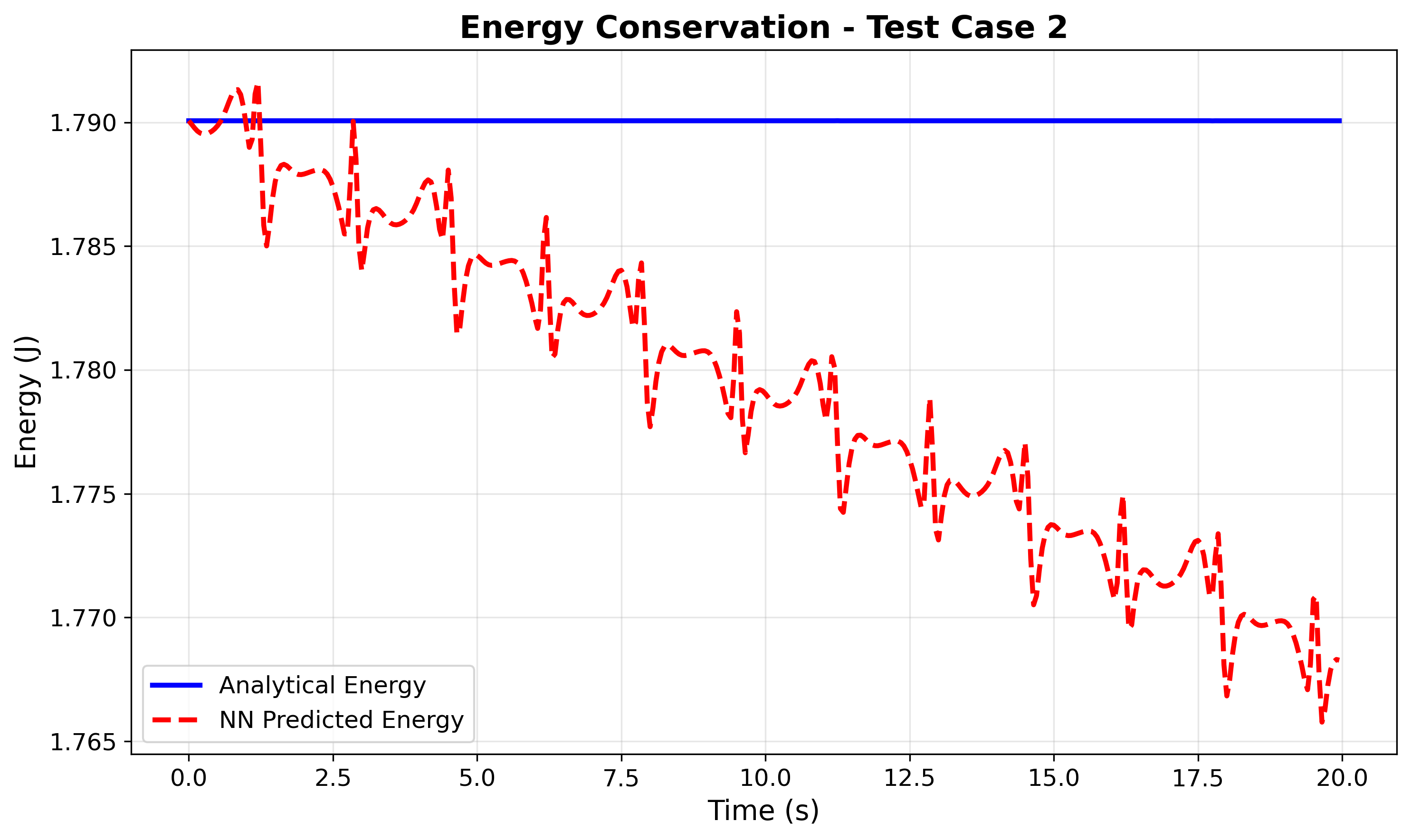}
\caption{\capSphereEnergy}
\label{fig:sphere_energy}
\end{figure}

\vspace{0.3cm}
\noindent\textbf{Lagrangian}

For geodesic motion on a sphere, the Lagrangian equals the kinetic energy (no potential):
\begin{align*}
L(\theta, \phi, \dot{\theta}, \dot{\phi}) &= \frac{1}{2} m R^2 \left( \dot{\theta}^2 + \sin^2\theta \, \dot{\phi}^2 \right)
\end{align*}

\noindent\textbf{Geodesic equations}
\begin{align*}
\ddot{\theta} &= \sin\theta \cos\theta \, \dot{\phi}^2 \\
\ddot{\phi} &= -2 \cot\theta \, \dot{\theta} \, \dot{\phi}
\end{align*}

\subsubsection*{Training Settings}

\squeezetable
\begin{ruledtabular}
\begin{tabular}{ll}
\textbf{Training data} & 50,000 uniformly random samples \\
\textbf{Position bounds} & $\theta \in [\pi/10, 9\pi/10]$, \\
& $\phi \in [-\pi/2, 5\pi/2]$ rad \\
\textbf{Velocity bounds} & $\dot{\theta}, \dot{\phi} \in [-5, 5]$ rad/s \\
\textbf{Validation data} & 10,000 (20\% of training) \\
\textbf{Epochs} & 50 \\
\textbf{Mini-batch size} & 128 \\
\textbf{Learning rate} & $10^{-3}$ \\
\textbf{Optimizer} & Adam with weight decay = $10^{-6}$ \\
\textbf{Scheduler} & Cosine Annealing, $T_{\max} = 30$, \\
& $\eta_{\min} = 10^{-6}$ \\
\end{tabular}
\end{ruledtabular}

\subsubsection*{Trajectory Test Settings}

\squeezetable
\begin{ruledtabular}
\begin{tabular}{ll}
\textbf{Initial conditions} & 4 uniformly random samples \\
\textbf{Position bounds} & $\theta \in [\pi/4, 3\pi/4]$ rad, $\phi \in [0, 2\pi]$ rad \\
\textbf{Velocity bounds} & $\dot{\theta}, \dot{\phi} \in [-2, 2]$ rad/s \\
\textbf{Time span} & $t \in [0, 20]$ s with $\Delta t = 0.05$ s \\
\textbf{Baseline solver} & LSODA~\cite{petzold1983} \\
& (rtol=$10^{-10}$, atol=$10^{-10}$) \\
\textbf{LNN solver} & PIDController~\cite{lienen2022} (rtol=$10^{-6}$, \\
& atol=$10^{-8}$, $p_{\text{coeff}}=0.2$, \\
& $i_{\text{coeff}}=0.4$, $d_{\text{coeff}}=0$) \\
\end{tabular}
\end{ruledtabular}

\subsubsection*{Network Architecture}

\squeezetable
\begin{ruledtabular}
\begin{tabular}{ll}
\textbf{Type} & MLP $+ 1 \cdot \|\dot{q}\|^2$ \\
\textbf{Layers} & 4 \\
\textbf{Hidden dimension} & 500 \\
\textbf{Activation} & GeLU \\
\textbf{Regularization} & Eigenvalue, $\lambda = 5$ \\
\end{tabular}
\end{ruledtabular}

\FloatBarrier

\subsection{AdS\textsubscript{4} Geodesic}

\begin{figure}[htbp]
\centering
\includegraphics[width=0.48\textwidth]{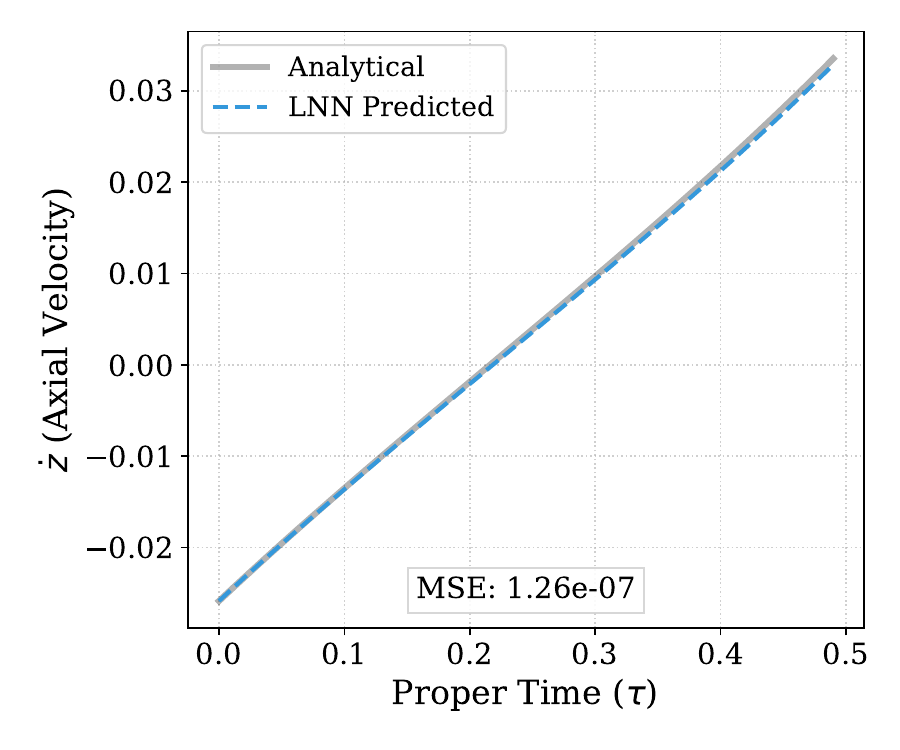}
\caption{\capAdSZDot}
\label{fig:ads4_zdot}
\end{figure}

\subsubsection*{Physical Settings}

\squeezetable
\begin{ruledtabular}
\begin{tabular}{ll}
\textbf{Inputs} & $[t, x, y, z, \dot{t}, \dot{x}, \dot{y}, \dot{z}]$ \\
\textbf{Physical scaling} & $s_t = 6.18$, $s_x = s_y = 5.00$, $s_z = 6.00$ \\
\textbf{Constants} & $L = 1$ (AdS radius) \\
\end{tabular}
\end{ruledtabular}

\vspace{0.3cm}
\noindent\textbf{Metric and Lagrangian}

We consider geodesic motion in four-dimensional Anti-de Sitter space (AdS$_4$) using Poincar\'{e} coordinates $(t, x, y, z)$ with $z > 0$. The metric is
\begin{equation}
    ds^2 = \frac{L^2}{z^2}\left(-dt^2 + dx^2 + dy^2 + dz^2\right).
\end{equation}
where $L = 1$ is the AdS radius. In these coordinates, the Lagrangian for a massive particle is
\begin{equation}
    \mathcal{L} = \frac{L^2}{2z^2}\left(-\dot{t}^2 + \dot{x}^2 + \dot{y}^2 + \dot{z}^2\right),
\end{equation}
where $\dot{} \equiv d/d\tau$ denotes differentiation with respect to proper time $\tau$. The geodesic equations are:
\begin{align}
    \ddot{t} &= \frac{2\dot{t}\dot{z}}{z}, \quad
    \ddot{x} = \frac{2\dot{x}\dot{z}}{z}, \quad
    \ddot{y} = \frac{2\dot{y}\dot{z}}{z}, \nonumber \\
    \ddot{z} &= \frac{\dot{t}^2 - \dot{x}^2 - \dot{y}^2 + \dot{z}^2}{z}.
\end{align}
For timelike geodesics, the constraint $g_{\mu\nu}\dot{x}^\mu\dot{x}^\nu = -1$ implies
\begin{equation}
    \dot{t} = \sqrt{\dot{x}^2 + \dot{y}^2 + \dot{z}^2 + \frac{z^2}{L^2}},
\end{equation}
so the four velocities $(\dot{t}, \dot{x}, \dot{y}, \dot{z})$ are not all independent.

\subsubsection*{Training Settings}

\squeezetable
\begin{ruledtabular}
\begin{tabular}{ll}
\textbf{Training data} & 120,000 random samples \\
\textbf{Position bounds} & $t, x, y \in [-5, 5]$, $z \in [0.5, 6]$ \\
\textbf{Velocity bounds} & $\dot{t} = 0$ (constrained), $\dot{x}, \dot{y} \in [-1, 1]$, \\
& $\dot{z} \in [-0.6, 0.6]$ \\
\textbf{Validation data} & 24,000 (20\% of training) \\
\textbf{Epochs} & 300 \\
\textbf{Mini-batch size} & 128 \\
\textbf{Learning rate} & $10^{-3}$ \\
\textbf{Optimizer} & Adam with weight decay = $10^{-6}$ \\
\textbf{Scheduler} & Cosine Annealing, $T_{\max} = 30$, \\
& $\eta_{\min} = 10^{-7}$ \\
\end{tabular}
\end{ruledtabular}

\subsubsection*{Trajectory Test Settings}

\squeezetable
\begin{ruledtabular}
\begin{tabular}{ll}
\textbf{Initial conditions} & 4 uniformly random samples \\
\textbf{Position bounds} & $t = 0$, $x, y \in [-1, 1]$, $z \in [0.5, 3]$ \\
\textbf{Velocity bounds} & $\dot{t} = 0$ (constrained), \\
& $\dot{x}, \dot{y}, \dot{z} \in [-0.25, 0.25]$ \\
\textbf{Proper time span} & $\tau \in [0, 0.5]$ with $\Delta \tau = 0.01$ \\
\textbf{Baseline solver} & LSODA~\cite{petzold1983} \\
& (rtol=$10^{-10}$, atol=$10^{-10}$) \\
\textbf{LNN solver} & PIDController~\cite{lienen2022} (rtol=$10^{-6}$, \\
& atol=$10^{-8}$, $p_{\text{coeff}}=0.2$, \\
& $i_{\text{coeff}}=0.4$, $d_{\text{coeff}}=0$) \\
\end{tabular}
\end{ruledtabular}

\subsubsection*{Network Architecture}

\squeezetable
\begin{ruledtabular}
\begin{tabular}{ll}
\textbf{Type} & MLP $+ 1 \cdot \|\dot{q}\|^2$ \\
\textbf{Layers} & 4 \\
\textbf{Hidden dimension} & 500 \\
\textbf{Activation} & GeLU \\
\textbf{Regularization} & Eigenvalue, $\lambda = 5$ \\
\end{tabular}
\end{ruledtabular}






\end{document}